%% file: main.tex
\PassOptionsToPackage{usenames,dvipsnames,table}{xcolor}
\documentclass[letterpaper, 10 pt, conference]{ieeeconf}  %

\IEEEoverridecommandlockouts                              %

\usepackage{times}

\usepackage{soul,xcolor}
\usepackage{multicol}
\usepackage{graphicx}

\usepackage{amsmath,amsthm,amssymb}
\usepackage[noend]{algpseudocode}
\usepackage{amsfonts}
\usepackage{amsmath}
\usepackage{float}
\usepackage{mathrsfs}
\usepackage{booktabs}

\let\labelindent\relax

\usepackage{enumitem}
\usepackage{algorithm}
\usepackage{algpseudocode}
\usepackage{todonotes}
\usepackage{multirow}
\usepackage{stfloats}
\usepackage{etoolbox}
\usepackage{caption}
\usepackage{capt-of}
\usepackage{duckuments}
\usepackage{makecell}

\usepackage[usenames,dvipsnames,table]{xcolor}
\usepackage[bookmarks=true]{hyperref}
\hypersetup{
    colorlinks=true,
    allcolors=black,
}
\usepackage[capitalise]{cleveref}

\title{\LARGE \bf
Whole-Body Dexterity
\vspace{-2mm}
}

\begin{document}
\author{
{\centering \bf
Conor Mc Gartoll$^{*1,2}$, An Dang$^{*1,3}$, Pranay Thangeda$^{*1}$, Arjun Gupta$^{1}$, Carolina Higuera$^{1}$,}\\
{\centering \bf
Nima Fazeli$^{1,3}$, Mustafa Mukadam$^{\dagger1}$,  Manikantan Nambi$^{\dagger1}$}
\\[2mm]
$^{1}$Amazon, $^{2}$University of California, San Diego, $^{3}$University of Michigan\\
$^{*}$Equal contribution, $^{\dagger}$Equal advising\\ 
\text{
\href{https://wholebodydex.com}{\textcolor{blue}{\nolinkurl{wholebodydex.com}}}
}
}

\IEEEaftertitletext{%
    \vspace{-2\baselineskip}%
    \begin{center}%
        \vspace{2mm}
        \includegraphics[width=\textwidth]{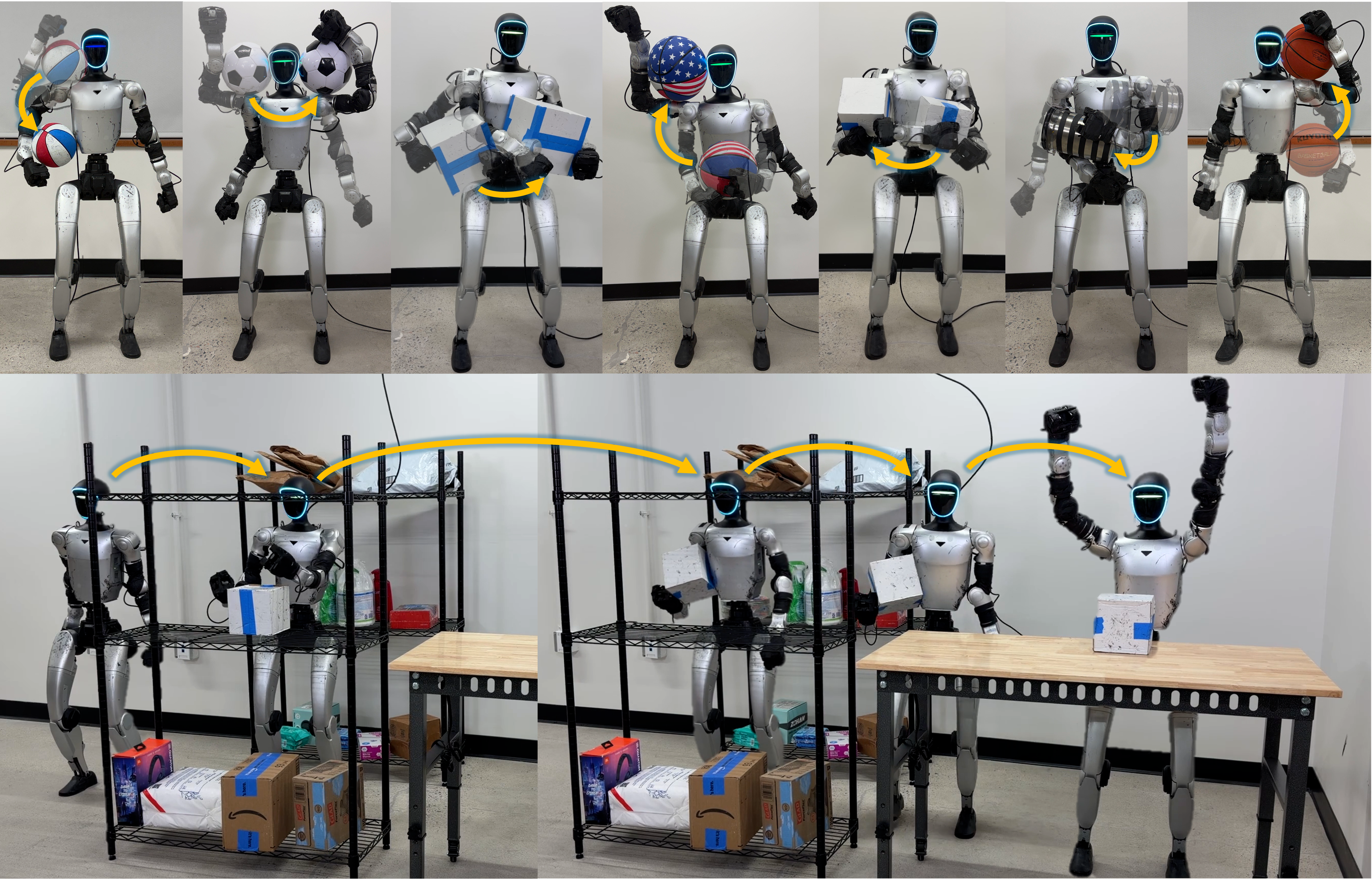}%
        \vspace{-1mm}
        \captionof{figure}{\textbf{Whole-Body Dexterity (WBD)} extends humanoid dexterous manipulation beyond the hands, allowing humanoids to mirror humans’ ability to use their arms and torso as contact surfaces for holding and repositioning objects. This enables human-like behavior such as lifting a box, securing it under one arm, carrying it while walking, and setting it down.}%
        \label{fig:teaser}
        \vspace{-1mm}
    \end{center}%
}

\maketitle
\thispagestyle{empty}
\pagestyle{empty}

\begin{abstract}

Current humanoid dexterous manipulation remains largely confined to end-effectors and simple grasps, falling short of humans’ ability to engage the entire body by bracing and sliding bulky objects against the arms and torso. We introduce Whole-Body Dexterity (WBD), a general framework that uses the robot's body and appendages as contact surfaces to grasp and manipulate objects. Our sim-to-real pipeline first generates a dataset of feasible whole-body grasps, then trains reinforcement-learning experts on a general any-to-any repositioning objective by sampling initial and goal grasps from the dataset, and distills these experts into a single policy that uses proprioception and object geometry. We show that the distilled policy transfers zero-shot to a real Unitree G1, where it repositions a basketball and a cardboard box to regions around the torso and completes a long-horizon task that stows a box under one arm, carries it from a shelf to a table, and places it. The controller operates interactively, steered by coarse motion references from a human teleoperator. Across real-world whole-body manipulation tasks, the distilled controller completes 90\% of trials, compared with 18\% for direct retargeting and 4\% for a generalist whole-body controller.

\end{abstract}

\input{text/01-introduction-v2}

\vspace{-1mm}
\section{Related Work}
\input{lit_review}

\section{Methodology}

We present a pipeline for whole-body dexterous manipulation, enabling object repositioning across diverse regions of the body. Beyond end-effector and front-of-torso contact, we define contact surfaces for grasping as any link or link subsection in the robot's kinematic system. Because the embodiment for whole-body grasping differs greatly from a multi-fingered hand, much of the re-grasping in the whole-body formulation involves sliding and rolling across the torso mesh, which we account for later in this section.

Our method trains and deploys controllers with dexterous whole-body re-grasping capabilities in three stages. We first generate a large dataset of whole-body object grasps using energy-based optimization and decompose our torso mesh into subregions to enable diverse whole-body grasping on the robot. Second, we train per-object experts on this grasp corpus to reposition objects around the body, encouraging sliding and rolling contact across the torso. Finally, we distill these experts into deployable controllers without privileged object state, enabling real-world whole-body dexterous object repositioning.

\subsection{Whole-Body Grasp Generation}
\label{sec:grasp-gen}

Given a kinematic system, we enumerate \emph{configuration types} (\cref{fig:grasp_gen}): which bodies or appendages hold the object and which of their links or subregions should contact it.
A \emph{grasp mode} $m \in \mathcal{M}$ abstracts a configuration type to the appendages contacting the object and is the level at which goals are commanded in \cref{sec:rl-training}. The system model, an object mesh, and the choice of contact links are the only inputs, so the procedure is not specific to a humanoid or to a class of objects.

A large body such as the G1's torso is contactable from every side, so optimizing against it as one contact target both concentrates contacts wherever they are easiest to find and collapses the distinct directions from which the object can be braced. We therefore subdivide large links into separate contact targets and generate grasps against each independently, which spreads the resulting grasps over the full extent of the body. On the G1, this subdivision yields named torso regions (\cref{fig:app-grasp-regions}), which the downstream training reuses to name goals, but the mechanism is generic. We showcase a variety of whole-body grasps across arms, legs, and the torso in \cref{fig:grasp_gen}. Leg grasps would require learning manipulation jointly with balance and locomotion, so we use only arm and torso grasps for WBD-EX training, giving the mode set $\mathcal{M}=\{L,B,R\}$: left-arm, both-arm, and right-arm grasps. In a single-arm mode, the other arm must release at the goal, and in mode $B$, both arms must contact it.

After defining our desired contact links and body subregions, we minimize the differentiable force-closure residual~\cite{liu2022synthesizingdiversephysicallystable} over the object's signed distance field, drawing contacts onto the object surface and keeping links out of the object and each other. Candidates are initialized from randomized object placements and optimized in GPU batches as done in large-scale dexterous hand grasp dataset generation~\cite{wang2023dexgraspnetlargescaleroboticdexterous, chen2025bodexscalableefficientrobotic, ye2025dex1blearning1bdemonstrations}. 

Because the G1 has only two arms and a large torso to grasp against, its grasps depend heavily on gravity and contact-link friction. Therefore, to produce the highest-quality grasp dataset, we aggressively filter candidates in simulation. For each candidate, the object is pressed into contact and held under $4\times$ gravity while a sustained wrench disturbance is swept over the six axis directions. The candidate passes only if contact persists. Survivors are then evaluated again under standard gravity without disturbance and must drift less than $10\,\mathrm{cm}$ and rotate less than $45^\circ$, producing the settled pose used at WBD-EX episode initialization.
We index accepted grasps by robot joint position, grasp mode, and settled object position, and drop grasps whose object mesh sits farther than $5\,\mathrm{mm}$ from the torso surface or below a size-dependent height threshold (\cref{app:cache-construction}). This concentrates the set on stable holds across the body, giving WBD-EX higher-quality grasps to transition between.

\subsection{WBD-EX Training}
\label{sec:rl-training}

We train WBD-EX to continually transition between object positions and grasp goals using proximal policy optimization (PPO)~\cite{schulman2017ppo}. We formulate this problem as a Goal-Conditioned Markov Decision Process (gcMDP) defined by the tuple $(\mathcal{S}, G, \mathcal{A}, \mathcal{R}, \mathcal{P}, \rho_0, \gamma)$, where $\mathcal{S}$ is the state space, $G$ is the goal set, $\mathcal{A}$ is the action space, $\mathcal{R}$ is the reward function, $\mathcal{P}$ is the transition probability, $\rho_0$ is the initial state distribution, and $\gamma$ is the discount factor. The robot seeks a policy $\pi_\theta(\mathbf{a} \mid \mathbf{s}, \mathbf{g})$, parameterized by $\theta$, that maximizes the expected discounted return $\mathcal{J} = \mathbb{E}_{\pi_\theta}\big[ \sum_{t=0}^{\infty} \gamma^{t} r_t \big]$ over control steps $t$, where $\mathbf{s} \in \mathcal{S}$, $\mathbf{g} \in G$, $\mathbf{a} \in \mathcal{A}$, and $r_t = \mathcal{R}(\mathbf{s}_t, \mathbf{a}_t, \mathbf{g})$.

Given the grasp dataset $\mathcal{D} = \{(q_i,\mathbf{X}_i^O)\}_{i=1}^{N_g}$ for $N_g$ grasps of an object, where $q_i$ is the settled robot joint configuration and $\mathbf{X}_i^O$ is the settled object pose (position and orientation) expressed in the torso frame, we voxelize grasps at 2\,cm resolution by settled object position and group each voxel's grasps by grasp mode. Each occupied (voxel, mode) pair is a node $n \in \mathcal{V}$ of a goal graph, and $\mathcal{D}(n) \subseteq \mathcal{D}$ denotes the grasps in that voxel with that mode. A transition between nodes is legal only if their modes are compatible: $L$ and $R$ each connect to themselves and to $B$, but never to each other, so every handoff passes through a both-arm hold.

We initialize each episode from the grasp dataset by sampling a mode uniformly, then a voxel carrying that mode uniformly, then \mbox{$g^\text{start} \sim \mathcal{U}(\mathcal{D}(n^\text{start}))$}, where $\mathcal{U}$ denotes the uniform distribution. Goals are commanded toward a \emph{far goal} whose mode, then voxel, are drawn uniformly. Each commanded goal advances toward it: we draw a goal distance $d \sim \mathcal{U}(0.02, d_{\max})$ in meters and command the legal-mode node within $d$ of the current node that is closest to the far goal, excluding the current node, with \mbox{$g^\text{goal} \sim \mathcal{U}(\mathcal{D}(n^\text{goal}))$}. A new far goal is drawn on arrival. With probability $0.2$, the command is instead a \emph{hold}, which asks the policy to keep the object at its current position and mode. To encourage human-like repositioning, in which objects roll and slide across body regions between arm handoffs, success depends only on the object's position, its grasp mode, and the object coming to rest there. Unlike~\cite{yin2025dexteritygenfoundationcontrollerunprecedented}, our goals and rewards exclude the goal joint configuration and object orientation.

We apply a goal-distance curriculum by ramping the ceiling $d_{\max}$ from $2$\,cm to $8$\,cm. Because each goal distance is drawn from $[2\,\text{cm}, d_{\max}]$, increasing $d_{\max}$ introduces progressively farther goal transitions while retaining closer ones. This exposes the policy to both gradual and more aggressive object repositioning behaviors as training progresses.

A goal is completed only when the object settles at the goal under the correct grasp mode (\cref{app:wbdex-goals}). For a single-arm goal, the inactive arm must register no object contact at the goal position. Requiring the inactive arm to fully release pushes the policy to learn stable and complete arm-to-arm handoffs around the body. We impose object-contact requirements throughout, but permit brief loss of the required arm contact within a $0.48$\,s grace window so the policy can execute the fast, dynamic reconfigurations that whole-body re-grasping demands rather than being forced into purely quasi-static holds.

\noindent\textbf{Rewards.} A goal is specified only by an object \emph{position} and a target grasp \emph{mode} (L, B, or R), so the policy is free to reach the target however it can. 
\emph{Retention} $\rho_t \in \{0,1\}$ indicates that the object has not dropped and keeps the arm contacts required by its current or target grasp mode; $e_t$ is the distance to the commanded position; and the inactive arm is \emph{clear} when it has no object contact. The reward is a weighted sum grouped as:
\begin{equation*}
r = \underbrace{r_{\mathrm{prog}} + r_{\mathrm{settle}} + r_{\mathrm{succ}} + r_{\mathrm{hold}}}_{\text{goal}}
  + r_{\mathrm{neutral}}
  - r_{\mathrm{fail}}
  - r_{\mathrm{reg}}
\end{equation*}
Progress, hold quality, and arm neutrality all carry the factor $\rho_t$ and are therefore zero whenever retention is lost.

\begin{itemize}[leftmargin=*, itemsep=2pt, topsep=3pt, parsep=0pt]
\item \textbf{Progress} $r_{\mathrm{prog}} = \rho_t \, \mathrm{clip}\!\left(e_{t-1} - e_t,\, \pm 2\,\mathrm{cm}\right)$. The
per-step reduction in distance to the commanded position, rewarded only while the object is retained and clipped so that no single step dominates.

\item \textbf{Goal-region arrival} $r_{\mathrm{settle}}$. A small per-step reward on movement goals during the dwell period, requiring the object to be settled within a $2$\,cm goal region in the required grasp mode.

\item \textbf{Transition completion} $r_{\mathrm{succ}}$. A sparse success bonus for repositioning the object and settling it at the goal under the required grasp mode.

\item \textbf{Hold quality} $r_{\mathrm{hold}} = \rho_t \exp(-e_t / 2.5\,\mathrm{cm})$ on hold goals, which command the object to stay where it is. It rewards proximity to the hold position while the object is retained.

\item \textbf{Arm neutrality} $r_{\mathrm{neutral}} = \rho_t \exp\!\left(-\lVert q^{\mathrm{arm}}_t - q^{\mathrm{arm}}_{\mathrm{def}} \rVert / 0.4\right)$. Attracts the arms toward their default configuration while the object stays held, so an arm not needed for the current grasp returns to its rest pose.

\item \textbf{Failure penalty} $r_{\mathrm{fail}}$. A single terminal penalty for a drop, sustained loss of object contact, or another physical or numerical failure (\cref{app:wbdex-terminations}).

\item \textbf{Regularizers} $r_{\mathrm{reg}}$. Five penalties for unsupported motion, excessive object speed, large actions, changes between actions, and joint-target tracking errors (\cref{app:wbdex-rewards}).
\end{itemize}

\noindent\textbf{Domain randomization.} We train an individual expert per object over a diverse set of primitives (spheres, cubes, rectangular prisms, and cylinders). To make holding and transitions robust to unmodeled dynamics, we randomize object mass, center of mass, and contact friction, and apply random perturbation forces to the object (\cref{app:wbdex-dr}). This drives the policy toward grasps and handoffs that stay stable under a range of physical conditions.

\noindent\textbf{Expert evaluation.} We assess WBD-EX expert quality by \emph{regional route completion}, the fraction of trials that reach every goal of a route in order before failure or timeout, where each route specifies a sequence of grasp goals across body regions. Route definitions, completion requirements, and evaluation details are given in \cref{app:route-eval}. Expert evaluation omits external force perturbations but still applies object-mass, friction, and center-of-mass domain randomization.

\noindent\textbf{Two-stage WBD-EX training.} When trained from scratch, experts for smaller and less uniformly shaped objects achieve lower regional route completion than experts for medium-large spheres. To reduce the time needed to train a WBD-EX expert for a new object, we pre-train medium-large sphere experts, which provide an effective initialization for the other primitives. We then warm-start additional object primitives from these pre-trained experts with an identical reward formulation. Warm-started experts achieve comparable or higher regional route completion than from-scratch experts at lower amortized GPU cost by sharing the faster sphere pre-training across target objects (\cref{app:warmstart}).

\subsection{WBD Controller Distillation}
\label{sec:refinement}

WBD-EX policies generate diverse and continuous object repositioning trajectories covering contactable regions of the torso, head, and arms under randomized dynamics.
We filter the trajectories to keep only valid repositioning trajectories, then balance sampling across objects and motion types with cross-body mirroring. We also sample three trajectory speed behaviors equally, using goal distances of 2--4, 4--6, and 6--8\,cm that are fixed within each trajectory, to capture both slow, deliberate and faster, dynamic repositioning behavior. See \cref{app:distill} for additional details.

These trajectories are used to train a WBD controller combining a conditional diffusion model with an inverse-dynamics network~\cite{yin2025dexteritygenfoundationcontrollerunprecedented}. Given proprioceptive history and a five-dimensional geometry descriptor comprising log-sorted bounding-box side lengths, log volume, and sphericity, it generates keypoint motions and converts them into joint-target increments. For real objects, we compute this descriptor from measured dimensions.
At inference, coarse commands guide generation through \textit{gradient guidance}~\cite{yin2025dexteritygenfoundationcontrollerunprecedented}. A tunable guidance weight controls how strongly denoising is steered toward the coarse motion command: larger weights emphasize command tracking, while smaller weights give greater relative influence to the learned motion prior. At each update, the diffusion model samples several candidate motions in parallel; we select the one closest to the command and convert it to refined joint commands with the inverse-dynamics model. This requires no privileged per-step object state and supports human teleoperation with tunable guidance.

\begin{figure*}[t]
    \centering
    \includegraphics[width=\textwidth]{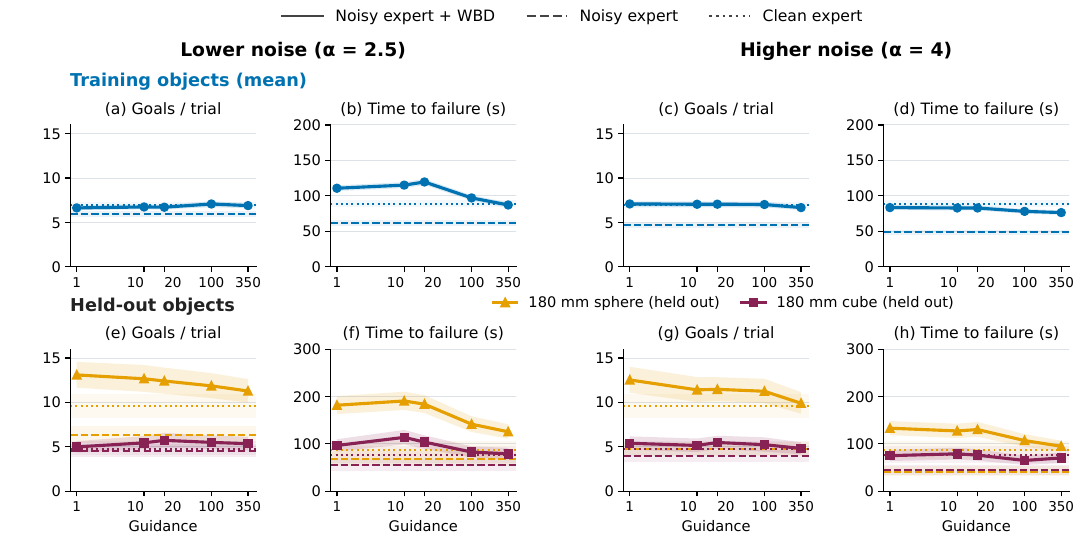}
    
    \vspace{-3mm}
    \caption{\textbf{Simulated object repositioning under noisy commands.}
    Trials sample transfers among six body regions (\cref{sec:sim-experiments}). Training-object mean (top) and held-out sizes (bottom). Solid, dashed, and dotted lines show WBD refinement averaged over three training seeds, noisy experts, and clean experts, respectively. Bands denote pointwise $95\%$ paired bootstrap confidence intervals clustered by source grasp over $256$ matched trials per object, with training objects weighted equally. Trials are capped at $600$\,s.}
    \label{fig:sim_results}
    \vspace{-6mm}
\end{figure*}

\input{text/04_experiments_results}

\section{Conclusion}

We presented a pipeline for learning whole-body dexterous manipulation at scale: contact configurations are enumerated into mode-typed whole-body grasps, and those grasps seed large-scale reinforcement learning under common rewards to dexterously reposition objects around the body. We showed how to use this large-scale whole-body dexterous manipulation data to train a shared controller that refines a coarse command stream into stable control and supports teleoperation. Although WBD-EX is trained without a lower-body controller, the distilled policy deploys on a standing and walking G1, holding objects through leaning and walking during the shelf task. Future work can naturally extend the same formulation to leg contact: generating combinations of arm and leg grasps, training WBD-EX jointly with a whole-body controller across this expanded grasp dataset, and distilling it into a policy to enable behaviors such as bracing a heavy box on a knee while the arms reposition.

\bibliographystyle{IEEEtran}
\bibliography{references}

\appendices
\crefalias{section}{appendix}
\crefalias{subsection}{appendix}

\newpage
\input{appendix}

\end{document}

%% file: text/01-introduction-v2.tex
\section{Introduction}

People rarely manipulate with their hands alone. We brace a box against the chest, cradle a basket across both arms, or tuck an object under one arm to free the other. These actions use the limbs and torso as a single highly dexterous manipulator.
We call this ability to reposition objects by continually changing supporting body contacts \emph{whole-body dexterous manipulation} and apply it directly to humanoids.

Prior humanoid work has used this term for specific settings, such as coordinated whole-body control for manipulation with the hands~\cite{twist2, humdex}, while related loco-manipulation confines dexterity to the end-effectors~\cite{amo, falcon, ultra, viral2025, psi02026} or to a sustained arm grasp on a bulky object~\cite{embracingbulky, tact, dao2024boxlocomanipulation}. None treats it as a general capability in which contact is established and released with any link to move an object across diverse regions of the body.

\begin{figure*}[t]
    \centering
    \includegraphics[width=\textwidth]{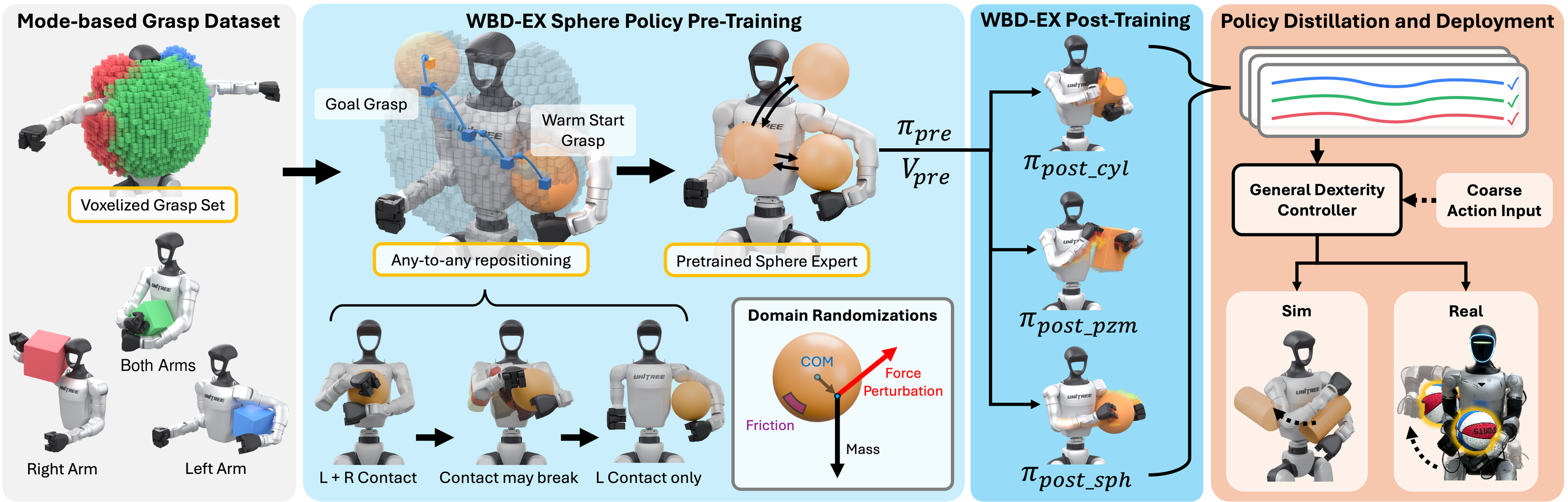}
    \caption{\textbf{Whole-Body Dexterity Pipeline.} \textbf{(1)} Generate a voxelized dataset of stable whole-body grasps. \textbf{(2)} Pre-train sphere WBD-EX experts on any-to-any repositioning using the grasp dataset. \textbf{(3)} Warm-start WBD-EX experts for other objects from the sphere experts. \textbf{(4)} Collect WBD-EX trajectories and distill them into a general dexterity controller for simulation and real-world tasks.}
    \label{fig:method}
    \vspace{-6mm}
\end{figure*}

Viewed through this lens, whole-body dexterous manipulation has a useful parallel in dexterous hands, with the limbs and torso taking the roles of fingers and palm. Recent dexterous-hand work shows that strong capabilities emerge from learning to move objects through dense sequences of intermediate targets, whether by pre-training on anygrasp-to-anygrasp transitions~\cite{yin2025dexteritygenfoundationcontrollerunprecedented} or by tracking hand--object subgoals from human motion through regrasps and finger gaiting~\cite{li2026teledexter}.
Extending this principle to humanoids introduces a challenge: a dexterous hand with many fingers can often reposition some fingers while the others support the object~\cite{hora,dactyl}, whereas a humanoid using two arms has fewer independent supports. Releasing one arm may require the other to brace the object against the torso, and changing grasps can require rapid reconfiguration and rolling or sliding contact. 
This additional complexity calls for a different approach to repositioning across body regions than these existing methods. 

We introduce \textbf{Whole-Body Dexterity (WBD)}, a framework for learning whole-body dexterous manipulation through \emph{mode-constrained re-grasping} (\cref{fig:method}) and deploying it on a real humanoid. We first generate a diverse set of whole-body grasps, then use them to initialize reinforcement-learning experts, \textbf{WBD-EX}. Experts learn to move objects between grasps toward goals specified by an object position and a target grasp mode: a left-arm, both-arm, or right-arm grasp. Transitions allow brief contact loss during movement but require a stable hold in the target mode at goals, so the experts learn how to hand off objects between arms, enabling repositioning across difficult regions such as behind the back and on top of the shoulders. 
We distill expert trajectories into a controller that uses an object geometry descriptor to refine motion commands into stable whole-body object repositioning, without privileged object state.
Our main contributions are:

\begin{enumerate}
    \item A whole-body grasp generation method defined over mode-typed contact configurations, formulated for arbitrary open-link articulated rigid-body systems and demonstrated on a humanoid robot.
    \item \textbf{WBD-EX}, per-object repositioning experts built on a mode-constrained re-grasping formulation that uses the diverse grasp dataset.
    \item A distillation procedure that turns these experts into a controller conditioned on proprioception and object geometry, refining coarse commands into dexterous manipulation of diverse objects in simulation and on a real humanoid. 
\end{enumerate}

 \begin{figure*}[t]
      \centering
      \includegraphics[width=\textwidth]{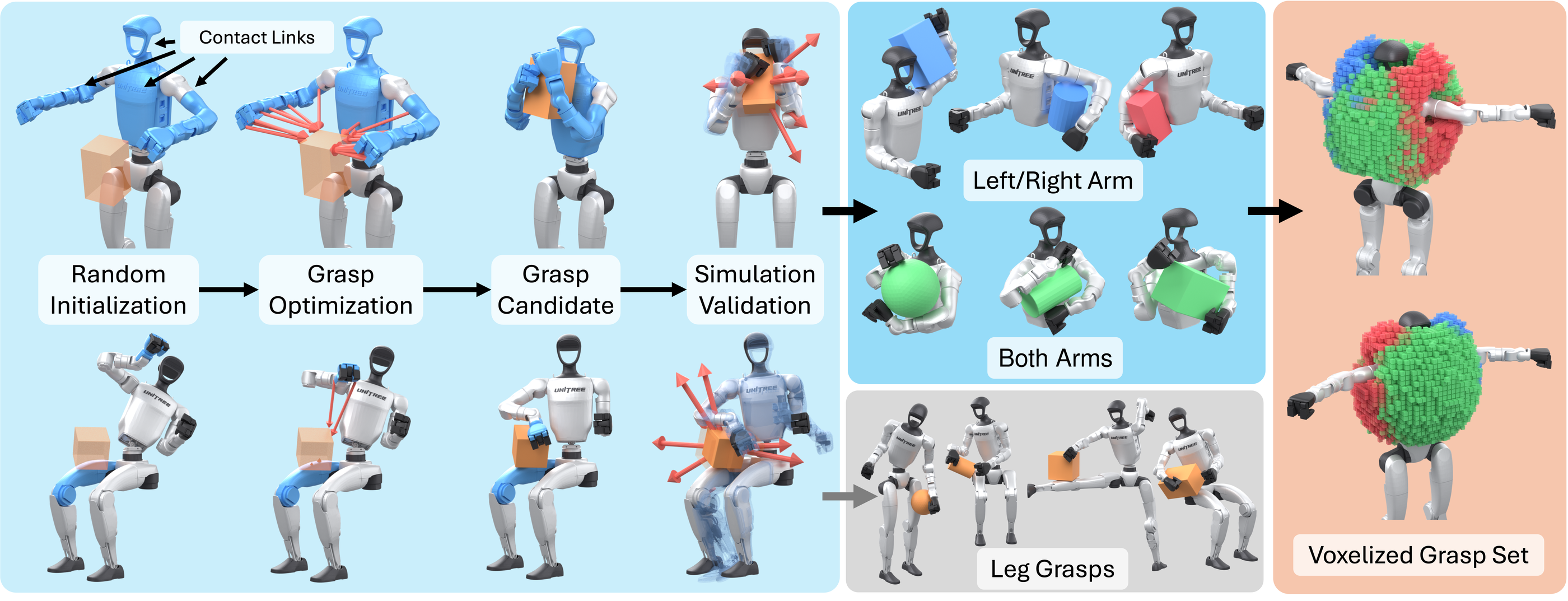}
      \vspace{-6mm}
      \caption{\textbf{Whole-Body Grasp Generation.} Energy-based optimization~\cite{liu2022synthesizingdiversephysicallystable, wang2023dexgraspnetlargescaleroboticdexterous, chen2025bodexscalableefficientrobotic} generates grasp candidates from specified arm, leg, and torso contact links (blue), followed by simulation validation under perturbations. The voxelized training set contains left-arm, right-arm, and both-arm grasps; leg grasps are excluded.}
      \label{fig:grasp_gen}
      \vspace{-6mm}   
\end{figure*}

%% file: lit_review.tex
\noindent\textbf{Whole-Body Manipulation.} In this paper, we advance whole-body dexterous manipulation, in which objects are held and repositioned through contact anywhere on the robot's body rather than only at designated end-effectors or links. The closely related term, whole-body dexterity, was defined for a snake-like robot arm that manipulates with its entire body surface~\cite{robopanoptes}. Related in spirit, a humanoid can manipulate
from onboard proprioception alone, since joint-encoder readings under compliant
contact act as a coarse whole-body tactile channel~\cite{blinddexterity}. Several methods manipulate bulky objects that exceed the hand's size and payload by
grasping them with the arms and torso and distributing contact across the
body~\cite{embracingbulky, punyo_pgrl, punyo_scirobotics, resmimic}. Other work uses upper-body tactile sensing to
hold boxes against the arms and chest on a real humanoid~\cite{tact}.
Closest to our setting,
reinforcement learning on a large soft robotic torso lifts cuboids of varying
sizes and masses and performs recovery re-grasping in front of the torso~\cite{softtorso}, but without standing or walking and
without moving the object to other regions of the body.

Classical model-based work
formulates this as optimization over contacts placed anywhere on the body
surface~\cite{explicitcontact, posturegen} and plans holding manipulation as a
transition graph over object motion and contact states~\cite{murooka2015}, though
typically under planar, quasi-static, or box-shaped assumptions; recent
learning-based methods instead expose whole-body contact as an explicit,
commandable interface~\cite{contactmimic} or imitate whole-body
human-object interactions in simulation from motion-capture demonstrations~\cite{intermimic, weave}. Evidence that the hand and the humanoid
form a structural continuum comes from reconfigurable hardware~\cite{handroid} and
from commercial humanoids that already tuck containers under one arm, drape cloth over a shoulder,
and lift appliances during everyday tasks~\cite{bd_atlas, figure_helix}.
Humanoid loco-manipulation couples locomotion with manipulation, enabling
dynamic behaviors such as climbing and wall flips alongside picking, placing, and pushing
objects~\cite{ultra, falcon, grail, motiondisco, omniretarget}, including humanoid pushing in simulation~\cite{sumo2026}.
We instead focus on general object manipulation across body regions.

Teleoperation is a direct route to whole-body manipulation and is well developed as motion transfer, through real-time human-to-humanoid retargeting and shadowing~\cite{he2024learningh20, omnih2o2024, humanplus2024, twist} and tracking controllers that follow whole-body targets~\cite{exbody2024, hover2025, amo, sonic2025}. However, without force or tactile feedback, it is difficult for the operator to tell whether an object is pinned or slipping, which largely limits teleoperation to hand-only pick-and-place behavior.

\vspace{2pt}
\noindent\textbf{Dexterity as Contact Sequencing.} In-hand manipulation can be framed as re-grasping: an object is carried through
a sequence of grasps by repeatedly breaking and remaking finger contacts, a process
known as finger gaiting~\cite{dactyl, hora, visualdexterity}. 
Using this idea, prior work trains dexterous-hand controllers that high-level policies or teleoperators steer with motion commands~\cite{yin2025dexteritygenfoundationcontrollerunprecedented, li2026teledexter}.
Related goal-conditioned approaches learn grasping and reorientation
across procedurally generated tools~\cite{kedia2026simtoolrealobjectcentricpolicyzeroshot}
or pre-train object reposing before adapting to downstream manipulation
tasks~\cite{lee2026adeptacceleratingdexteritypretraining}. Similar in mechanism to our own
approach, a constrained sampler over feasible contact-rich states can guide goal-conditioned
manipulation RL directly~\cite{26-toussaint-CSRL}. We lift this re-grasping principle from dexterous hands and arms to the whole humanoid.

%% file: text/04_experiments_results.tex
\section{Experiments and Results}

We evaluate whether WBD can be used to generate whole-body dexterous data and train policies that refine imperfect motion commands into stable whole-body dexterous manipulation. We first report grasp and WBD-EX data generation, then evaluate a distilled WBD controller for on-body repositioning in simulation and on a real Unitree G1. A shelf pick, stow, carry, and place task tests whether these skills compose into a longer manipulation sequence.

\subsection{Experimental Setup}
Our object set contains 15 primitives: six spheres, four cubes, three rectangular prisms, and two cylinders. All side lengths and diameters fall between 80 and 240\,mm. WBD-EX uses 2.51
million simulation-stable grasps passing the dataset filters (\cref{sec:grasp-gen}). Mean
per-object coverage of sampled object-center positions around the torso is 90.1\% using each object's minimum bounding-box half-extent as the distance tolerance; pooled coverage is 98.4\%
using the mean minimum half-extent (\cref{app:grasp-coverage}).

We train one WBD-EX policy per object using the procedure in \cref{sec:rl-training} and distill a shared WBD controller from their trajectories. Expert proficiency varies with object geometry: sphere experts achieve the highest overall route completion, while experts for cubes, prisms, and cylinders still reposition across all body regions but with lower overall success rates (\cref{app:expert-results}). Because distillation uses only trajectory segments that pass a quality filter (\cref{app:distill-data}), less proficient experts yield valid demonstrations less efficiently, requiring more collection attempts per accepted trajectory. Our training dataset consists of 274 thousand trajectories and 873 million recorded frames; training samples objects equally. Network architectures, optimization settings, data augmentation, and training costs are detailed in \cref{app:wbdex,app:distill-optimization,app:warmstart-cost}.

\subsection{Simulation Experiments}
\label{sec:sim-experiments}

We evaluate on-body repositioning in simulation using a WBD controller trained on 13 of the 15 primitives (three training seeds), with a 180\,mm sphere and a 180\,mm cube withheld from training, validation, and normalization. We degrade each of the 15 objects' WBD-EX policies into coarse command streams and compare the degraded expert alone against the WBD controller refining it. We add independent uniform noise $\mathcal{U}(-\alpha,\alpha)$ to all 14 Unitree G1 arm commands. We report results at $\alpha\in\{2.5,4\}$ and guidance weights $\{1,10,20,100,350\}$, with clean experts as a reference (\cref{fig:sim_results}). WBD applies bounded corrections to the noisy commands, within $\pm2.5$ of each noisy command in the experts' pre-$\tanh$ action space (\cref{app:distill-inference})%
. Degrading an expert reproduces the failure we see with real operator commands: the stream still carries the goal direction while individual arm targets move far enough to break supporting contact mid-handoff. In both settings, the failure is a loss of supporting contact followed by a drop.

\begin{figure}[t]
    \centering
    \includegraphics[width=\columnwidth]{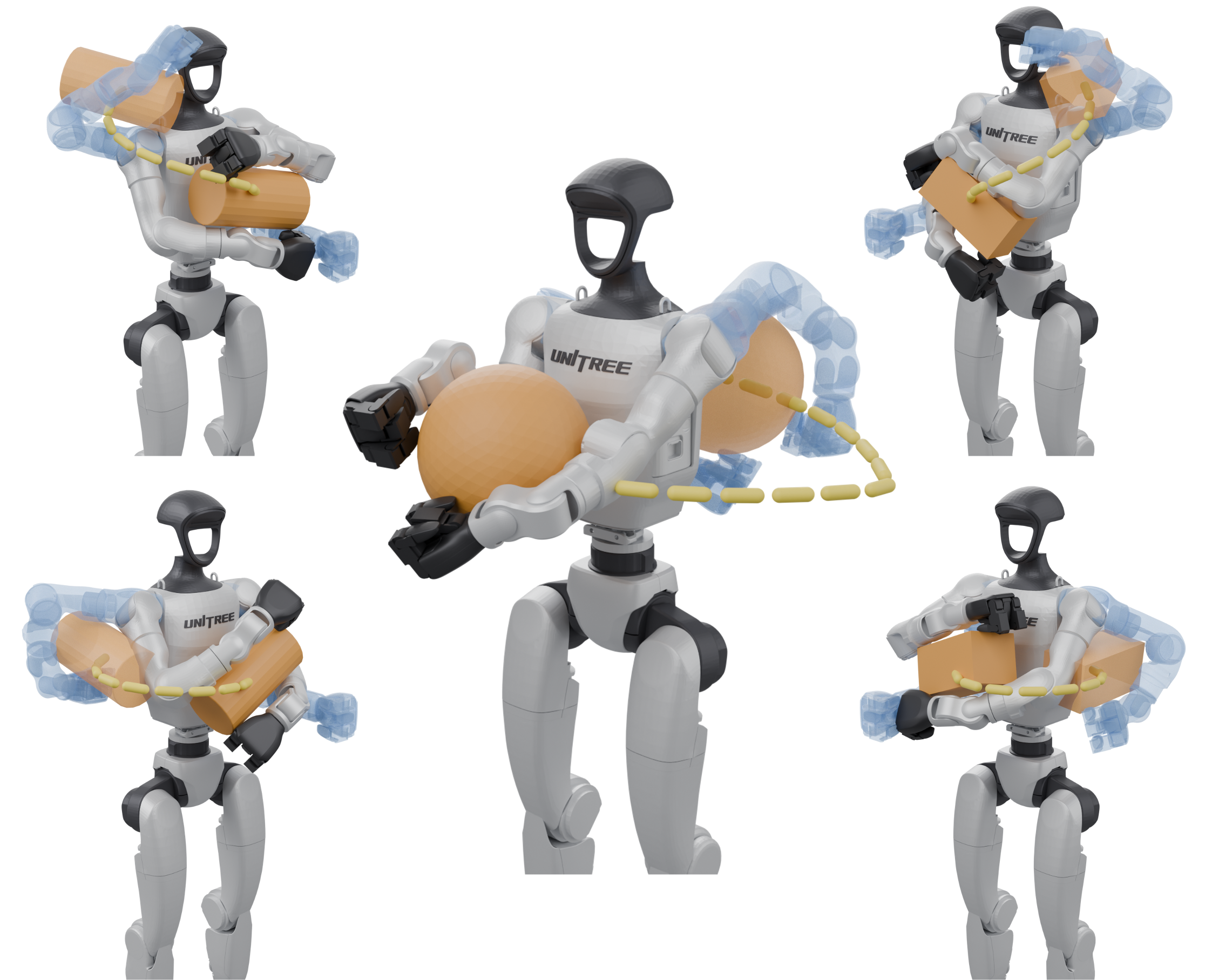}
    \vspace{-4mm}
    \caption{\textbf{Simulated WBD controller experiment.} Examples of regional transfers in which the start grasp is shown on the fully visible G1, while the sampled \textit{far goal} grasp in another region is shown as the blue ghost.}
    \vspace{-10mm}
    \label{fig:sim_fig}
\end{figure}
We evaluate transfers among six body regions (\cref{fig:sim_fig}): front, both underarms, both shoulders, and back. Each trial moves the object through a seeded random sequence of adjacent regions (\cref{app:sim-region-walk}). The WBD controller is trained across all regions and voxels around the body; see \cref{app:distill-sampling}. Each underarm and shoulder goal requires the opposite arm to release, with both arms supporting the object at the front and back. Each condition uses 256 matched initial grasps per object spanning the six regions; initial states and random seeds are matched across conditions. Trials end at the first failure or 600\,s. We report completed goals per trial and time to failure restricted to this horizon (\cref{fig:sim_results}). 
Each goal is a (voxel, grasp mode) pair sampled from the voxel goal graph along the path toward a randomly drawn far goal in the next region of the sequence; far goals are included as goals. 
Tracking the number of goals achieved is an effective measure of how well a policy can stably transition an object across regions of the body.

Despite noisy commands, WBD largely recovers the clean experts' mean goal-reaching performance across training objects at both plotted noise levels (\cref{fig:sim_results}). Averaged over these two noise levels and all five guidance values, mean time to failure increases relative to the noisy expert by 70\% on training objects and by 158\% and 71\% on the held-out sphere and cube, respectively. Training objects are weighted equally. At the lower noise level and guidance weights of 1--20, WBD even exceeds the clean experts' time to failure, by 25--35\% for the training-object mean and by 109--120\% and 25--47\% for the held-out sphere and cube, respectively, showing that the WBD controller maintains stable object holding beyond what the uncorrupted experts achieve.
As \cref{fig:sim_results} shows, excessive command tracking can reduce both stability and goals per trial.

\begin{figure*}[t]
    \centering
    \includegraphics[width=\textwidth]{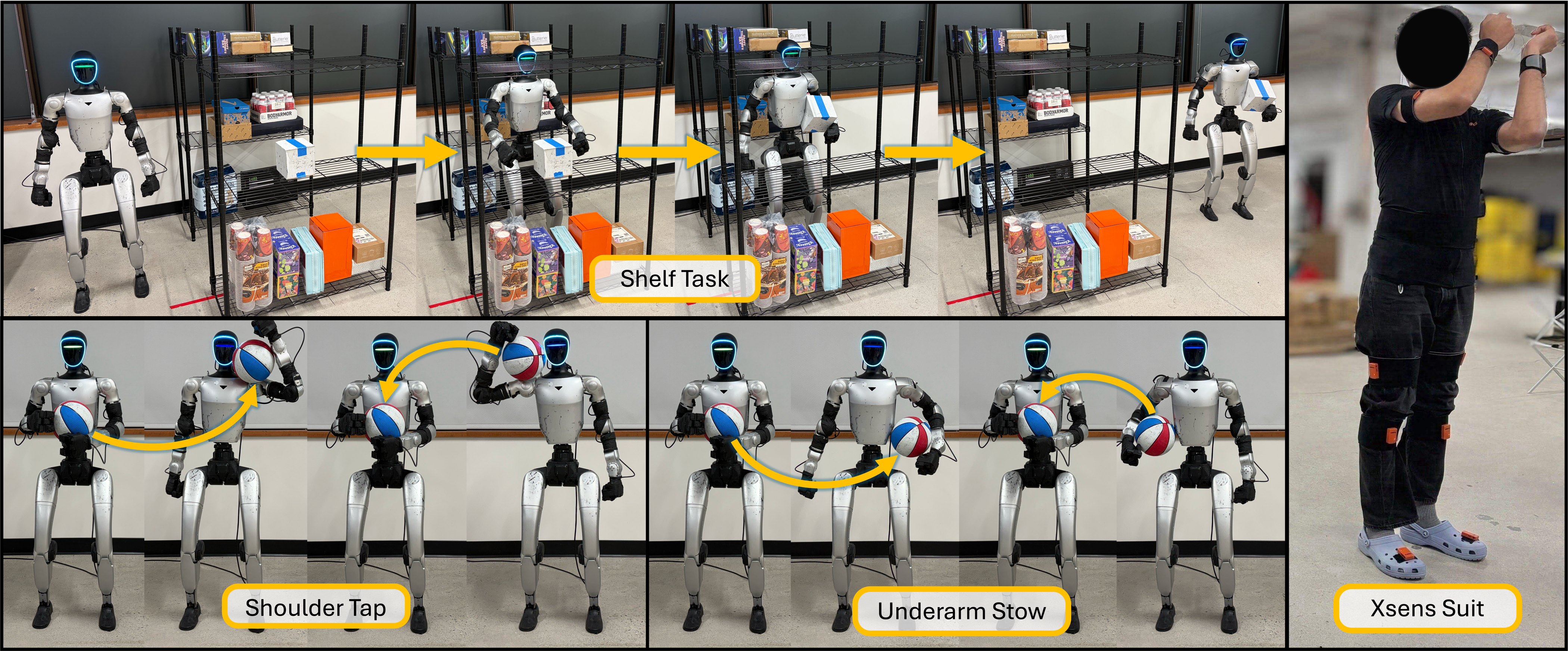}
    \caption{\textbf{Real-world whole-body dexterous manipulation.} On-body repositioning (bottom) and shelf pickup, stowing, and carrying (top) using the WBD controller, guided by the teleoperator (right).}
    \vspace{-6mm}
    \label{fig:auto_exp}
\end{figure*}

\subsection{Real-World Experiments}

We evaluate a WBD controller trained on the full 15-object WBD-EX trajectory dataset on two on-body repositioning tasks and a long-horizon shelf task (\cref{fig:auto_exp}). For repositioning, a size 3 basketball or a cardboard box (15\,cm cube) starts in a two-handed front grasp. The \textit{shoulder tap} task succeeds when the object contacts the shoulder link, and the \textit{underarm stow} task succeeds when the object is held stably under one arm, with its center approximately as far forward as the shoulder link. A round trip additionally requires returning to a stable two-arm front grasp. We report region-reaching success, full round-trip success, and mean time to each endpoint over successful trials (\cref{tab:real_results}). The shelf task requires picking the same cardboard box from a shelf, stowing it under one arm, stably carrying it while walking, un-stowing it, and placing it on a table. Pick and un-stow require a stable two-arm front grasp, and stow uses the underarm criterion, except that the free arm may stabilize the box while walking. We report success counts for each subtask: pick, stow, carry, un-stow, and place (\cref{tab:real_shelf}).

All trials are run by a single skilled human teleoperator with $8+$ hours of training on these tasks. We use an Xsens motion capture suit to track motion beside and behind the operator, where standard VR headset tracking is limited. The controller can also accept commands from VR headsets and trackers, although limited tracking coverage can restrict dexterity. The operator indicates the desired motion, while WBD refines the arm targets to maintain object support as the arms and torso reconfigure. For the long-horizon task, the operator switches the guidance weight between $10$ for on-body repositioning and $100$ for pick-and-place, balancing stable support with responsive command following.

We compare WBD refinement of motion retargeted by General Motion Retargeting (GMR)~\cite{araujo2025retargeting} against direct GMR control and SONIC~\cite{sonic2025}. For WBD and direct GMR, GMR retargets the Xsens motion to G1 upper-body targets, and Adaptive Motion Optimization (AMO)~\cite{amo} controls the lower body and balance. SONIC runs its public checkpoint in VR three-point mode, tracking Xsens head and wrist poses while its kinematic planner controls the lower body.
The real-world results suggest that maintaining object support through contact transitions is a central bottleneck (\cref{tab:real_results,tab:real_shelf}). GMR let the operator move the arms toward the relevant body regions, but stable contact during repositioning remained difficult; SONIC additionally struggled with reachability in constrained regions. Across the three tasks, counting complete on-body round trips for both objects and full shelf sequences, the WBD controller completed 45 of 50 trials (90\%), compared with 9 of 50 (18\%) for GMR and 2 of 50 (4\%) for SONIC.

In simulation, cube WBD-EX experts have lower mean regional route completion than sphere experts (\cref{tab:app-expert-results}). In the real-world shoulder tap trials, however, GMR completed $4/10$ box round trips versus $1/10$ with the basketball, with mean completion times of $8.5$ versus $30.1\,\mathrm{s}$ among successful trials. 
We hypothesize that the cardboard's lower friction makes it easier to slide to the shoulder, whereas the basketball's higher friction requires rolling along the torso and more deliberate arm manipulation, consistent with the faster successful WBD and GMR box trials. Underarm stowing still remained difficult, requiring a complete single-arm handoff, and yielded low success rates for GMR and SONIC ($1/10$ and $0/10$ complete box round trips, respectively, compared with $10/10$ for WBD).

\begin{table*}[t]
\centering
\footnotesize
\setlength{\tabcolsep}{3pt}
\renewcommand{\arraystretch}{1.1}
\begin{tabular*}{\textwidth}{@{\extracolsep{\fill}} ll*{8}{c}}
\toprule
\multirow{2}{*}{Task} & \multirow{2}{*}{Controller}
& \multicolumn{4}{c}{Basketball}
& \multicolumn{4}{c}{Cardboard box} \\
\cmidrule(lr){3-6}\cmidrule(lr){7-10}
& & Reached $\uparrow$ & Round trip $\uparrow$
& Region (s) $\downarrow$ & Total (s) $\downarrow$
& Reached $\uparrow$ & Round trip $\uparrow$
& Region (s) $\downarrow$ & Total (s) $\downarrow$ \\
\midrule
\multirow{3}{*}{Shoulder tap} & SONIC & 2/10 & 0/10 & 15.2 & ---
& 2/10 & 2/10 & 15.9 & 24.9 \\
& GMR & 3/10 & 1/10 & 26.0 & 30.1
& 6/10 & 4/10 & 8.0 & 8.5 \\
& WBD (ours) & \textbf{10/10} & \textbf{9/10} & \textbf{7.7} & \textbf{17.3}
& \textbf{10/10} & \textbf{10/10} & \textbf{4.1} & \textbf{6.3} \\
\cmidrule(l){1-10}
\multirow{3}{*}{Underarm stow} & SONIC & 1/10 & 0/10 & 28.0 & ---
& 2/10 & 0/10 & 26.1 & --- \\
& GMR & 4/10 & 2/10 & 31.5 & 49.8
& 2/10 & 1/10 & 9.3 & 17.8 \\
& WBD (ours) & \textbf{10/10} & \textbf{9/10} & \textbf{10.6} & \textbf{26.2}
& \textbf{10/10} & \textbf{10/10} & \textbf{8.6} & \textbf{13.0} \\
\bottomrule
\end{tabular*}
\caption{\textbf{Real-world on-body repositioning.} Ten trials per controller, object, and task, five per side, with a size 3 basketball and a cardboard box (15\,cm cube): region-reaching and full round-trip success. Times are measured from trial start and averaged over trials reaching the respective endpoint.}
\vspace{-6mm}
\label{tab:real_results}
\end{table*}

\begin{table}[h]
\vspace{-2mm}
\centering
\footnotesize
\setlength{\tabcolsep}{3pt}
\renewcommand{\arraystretch}{1.1}
\begin{tabular}{l*{5}{c}}
\toprule
Controller & Pick $\uparrow$ & Stow $\uparrow$ & Carry $\uparrow$ & Un-stow $\uparrow$ & Place $\uparrow$ \\
\midrule
SONIC & 10/10 & 0/10 & 0/10 & 0/10 & 0/10 \\
GMR & 10/10 & 3/10 & 2/10 & 1/10 & 1/10 \\
WBD (ours) & 10/10 & \textbf{9/10} & \textbf{9/10} & \textbf{9/10} & \textbf{7/10} \\
\bottomrule
\end{tabular}
\caption{\textbf{Real-world shelf task.} Completed stages over ten trials per controller with the cardboard box.}
\vspace{-2mm}
\label{tab:real_shelf}
\end{table}

On the long-horizon task (\cref{tab:real_shelf}), all methods completed the initial shelf pickup, but subsequent stowing and transport required support to persist as the robot reconfigured and moved. The WBD controller maintained support through carrying and un-stowing after successful stows, showing that the controller can stably hold objects while being perturbed. While substantially better than GMR and SONIC, WBD still lost contact in a few trials, dropping the object.

%% file: appendix.tex
\section{Whole-Body Grasp Generation Details}
\label{app:grasp-gen}

We adapt prior energy-based dexterous grasp synthesis~\cite{ye2025dex1blearning1bdemonstrations, wang2023dexgraspnetlargescaleroboticdexterous} to prescribed contact links on any combination of arms, legs, and torso, with the torso serving as a fixed support surface and kinematic reference.
We seek joint configurations $q \in \mathcal{C}$ in configuration space $\mathcal{C}$ that minimize the resulting objective. We define $K$ contact points $\mathbf{p}_j$  on the link mesh surfaces throughout the robot that we wish to make contact with the object and, using the signed distance function of the object mesh $\phi: \mathbb{R}^3 \rightarrow \mathbb{R}$, positive outside the object and negative inside, we formulate the following:
\begin{align}
    \min_{q, \mathbf{T}} \quad & E_{\text{closure}} + \frac{1}{K}\sum_{j=1}^K [\phi(\mathbf{p}_j) - d^{\star}]_+^2\\
    \text{s.t.} \quad & \phi(\mathbf{b}_i) \geq r_i, \quad \forall i = 1,\dots,N \\
    &\| \mathbf{b}_i - \mathbf{b}_l \|_2 \geq r_i + r_l, \quad \forall (i,l) \in \mathcal{H} \\
    & \underline{q}_m \leq q_m \leq \bar{q}_m, \quad \forall m = 1,\dots,n_q
\end{align}

where $E_{\text{closure}}$ is a force-closure energy defined below, $\mathbf{T} \in SE(3)$ is the pose of the robot's kinematic root in the object frame, $\underline{q}_m$ and $\bar{q}_m$ are the limits of joint $q_m$ for the $n_q$ optimized joints, $\mathbf{b}_i$ are the centers of the robot's $N$ collision spheres with radii $r_i$, and $\mathcal{H}$ is the set of non-adjacent sphere pairs checked for self-collision. The contact-distance tolerance $d^\star$ sets the distance beyond
which a contact point is attracted toward the surface. Here, $[x]_+ = \max\{x,0\}$ denotes the positive part of $x$: it returns $x$ when positive and zero otherwise. Each squared penalty therefore applies only when its corresponding threshold is violated. We constrain the optimization problem above to avoid collisions between the link collision spheres and the object, avoid self-collision between link collision spheres, and obey joint limits.

Since this problem is nonconvex and we would like to leverage the GPU parallelism of first-order solvers such as Adam to generate many grasps simultaneously, we relax this optimization problem to the unconstrained form:
\begin{align*}
    \min_{q, \mathbf{T}} \quad & w_{\text{closure}} E_{\text{closure}} + \frac{w_{\text{dist}}}{K}\sum_{j=1}^K [\phi(\mathbf{p}_j) - d^{\star}]_+^2 \\
    &+ w_{\text{link}}\sum_{i=1}^N [r_i - \phi(\mathbf{b}_i)]_+^2 \\
    &+ w_{\text{joint}}\sum_{m=1}^{n_q} \big([\underline{q}_m + \delta_q - q_m]_+^2 + [q_m - \bar{q}_m + \delta_q]_+^2\big) \\
    &+ w_{\text{self}}\sum_{(i,l) \in \mathcal{H}} [ r_i + r_l - \|\mathbf{b}_i - \mathbf{b}_l \|_2]_+^2
\end{align*}

where we define weights $w$ for each optimization term. The joint-limit penalty begins
$\delta_q$ inside each limit; joint angles are also clamped to their
hard limits after each update. \Cref{tab:app-grasp-parameters} lists the WBD grasp-generation settings.

We use the following differentiable force-closure energy, adopted from prior work on dexterous hand grasp synthesis:
\[
E_{\text{closure}} = \frac{1}{K^2} \left(
    \left\|\sum_{k=1}^K \mathbf{n}_k\right\|_2^2
    + \left\|\sum_{k=1}^K (\mathbf{p}_k - \bar{\mathbf{p}}) \times \mathbf{n}_k\right\|_2^2
\right),
\]
where $\mathbf{n}_k = -\nabla \phi(\mathbf{p}_k) / \| \nabla \phi(\mathbf{p}_k) \|_2$ is the inward-pointing unit surface normal at the object surface point closest to contact point $\mathbf{p}_k$, and $\bar{\mathbf{p}} = K^{-1}\sum_{k=1}^K \mathbf{p}_k$ is the contact centroid. $E_{\text{closure}}$ adapts the force-closure energy of~\cite{liu2022synthesizingdiversephysicallystable, wang2023dexgraspnetlargescaleroboticdexterous, ye2025dex1blearning1bdemonstrations}: it is the sum of the squared net force and
moment of unit normal contact forces, normalized by $K^2$, so it vanishes when both cancel. Contact magnitudes are fixed at unity and no friction cone enters, making this a
frictionless net-wrench residual rather than a force-closure feasibility test, and moments
are taken about the contact centroid rather than the object's center of mass. Surface normals are held fixed during each gradient update.

\subsection{Whole-Body Grasp Parameterization}

The humanoid is treated as a dexterous manipulator whose torso acts as a static
``palm.'' Any selected arms or legs can move to establish the prescribed contacts.
The joint vector $q$ contains the $n_q$ joints selected for optimization; all other
joints remain fixed at their prescribed configurations.

A candidate has $n_q+9$ scalars optimized jointly: the three-dimensional position of
the kinematic root in the object frame, its orientation in a continuous
six-dimensional rotation parameterization, and the $n_q$ joint angles. The signed
distance field is defined in the object frame with the object at the origin.
Optimizing $\mathbf{T}$ chooses the relative placement of the object and robot;
inverting it gives the object pose relative to the fixed torso.

\subsection{Geometry and Contact Sampling}

The object is represented by the exact triangle-mesh signed distance field queried
through a bounding-volume hierarchy; no voxel volume is used, so all iterations query
the mesh directly. Signs are computed by angle-weighted pseudonormals during the search
and by ray parity for the final refinement iterations.
The robot is approximated by collision spheres rather than by link meshes~\cite{ye2025dex1blearning1bdemonstrations},
generated from the robot geometry by spherical approximation~\cite{2025foamtoolsphericalapproximation}.

Candidate contact points are sampled on the link meshes separately from the collision
spheres. A contact configuration specifies the allowed arm, leg, and torso links and
which links are required. Each iteration samples $K$ allowed links, always
including the required links, and draws one candidate point on each, resampled every iteration~\cite{wang2023dexgraspnetlargescaleroboticdexterous}. This keeps contacts from
collapsing onto a single link and produces the spread of contact patterns the dataset
exhibits. Penetration is evaluated over all spheres, regardless of active contacts.

Contact selection can also restrict torso contact points to a named surface region.
These regions are tags on individual candidate points, rather than geometric boxes.
They restrict the points used in the contact objective, while collision checks
continue to cover the whole robot.

\subsection{WBD Grasp Dataset}

For WBD-EX training and the rest of the pipeline, we specialize this general method
to grasps using the arms and torso. On the Unitree G1, the Dex3 finger joints are
pinned to constant closed-grip values, and the twelve leg joints and three waist
joints remain fixed. Bimanual generation optimizes the fourteen arm joints, giving
$23$ scalar variables including the root pose; single-arm generation additionally
freezes the opposite arm, leaving seven optimized joints and $16$ scalar variables.
The numerical settings and validation procedure below describe this WBD grasp dataset.

The candidate contact set contains $626$ points, sampled uniformly on the link meshes
and then hand-curated, across $34$ links; $347$ points over $25$ links are active for
the bimanual configuration. We sample $K=5$ contact links per iteration.
The dataset uses three contact configurations: right-arm contact over $13$ links and
bimanual contact over $25$ links, both requiring torso contact, as well as bimanual
contact over $24$ arm links with the torso excluded.

\begin{figure}[t]
    \centering
    \includegraphics[width=\columnwidth]{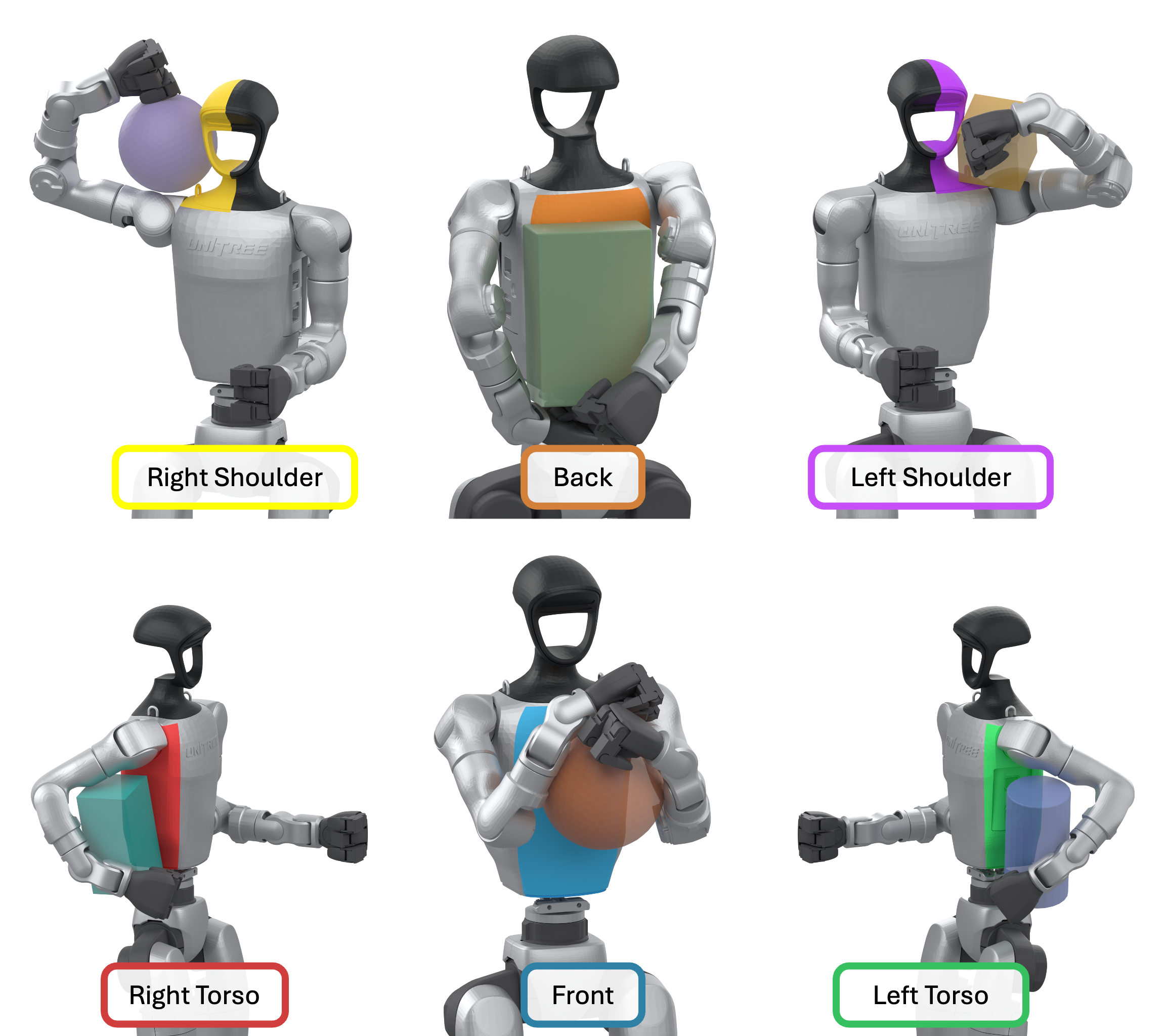}
    \caption{\textbf{Torso contact regions for the WBD grasp dataset.} One accepted
    arm--torso grasp is shown per region, each on a different object.}
    \label{fig:app-grasp-regions}
    \vspace{-4mm}
\end{figure}

For this dataset, only torso contact points carry region tags, spanning the six
regions shown in \Cref{fig:app-grasp-regions}.

Restricting to a region has four effects. It filters the candidate contact points, so a
generation run restricted to one torso region optimizes against roughly a sixth of the available torso points; it
biases initialization, since the direction the body is aimed from is set per region; it
assigns each region an equal share of the per-object grasp generation quota;
and it labels the resulting grasps so that region coverage can be
audited afterward. It does not restrict which links may make contact.

Every accepted grasp is finally mirrored across the sagittal plane, which doubles the
set exactly. The mirror is an exact
relabeling of an already-validated grasp, so it is not re-simulated.

\subsection{Pre-Grasp and Post-Grasp Targets}

We adapt the contact-normal pressing strategy of~\cite{wang2023dexgraspnetlargescaleroboticdexterous} for the post-grasp target and extend it to a retracted pre-grasp.
From a solved configuration, we derive a retracted pre-grasp and a pressed post-grasp
target by backpropagating $\sum_k \nabla\phi(\mathbf{p}_k)\cdot \mathbf{p}_k$ through the kinematics, with the SDF gradients held fixed, to obtain a
joint-space gradient. We normalize this gradient to unit Euclidean norm,
multiply it by $+0.5$ for retraction or $-0.4$ for pressing, and add it to $q$,
clamping the result to the joint limits. The pre-grasp is the
approach pose used to start the simulation trial; the post-grasp is the press target
held during validation and later reused as the servo setpoint at each WBD-EX episode
reset.

\subsection{Solver}

For the WBD grasp dataset, candidates are optimized in parallel on one GPU: $4096$ candidates for
$4000$ iterations with AdamW at learning rate $10^{-3}$, weight decay $10^{-4}$, and a
cosine schedule annealing to $10^{-4}$. Steps after a short warmup are replayed from
captured CUDA graphs. 
Optimization candidates are initialized by sampling a random direction outward from the object centroid and rotating the robot
to face the object centroid. Orientation is then jittered by a uniform roll about the
facing axis over $(-\pi,\pi)$ and two orthogonal tilts over $(-\pi/8,\pi/8)$, and the
seed posture is jittered per joint by $\pm 25\%$ of that joint's range.

\begin{table}[h]
\centering
\begingroup
\footnotesize
\setlength{\tabcolsep}{5pt}
\begin{tabular}{lr}
\toprule
Parameter & Value \\
\midrule
Closure weight $w_{\mathrm{closure}}$ & $1$ \\
Contact-distance weight $w_{\mathrm{dist}}$ & $300$ \\
Object-penetration weight $w_{\mathrm{link}}$ & $200$ \\
Joint-limit weight $w_{\mathrm{joint}}$ & $1$ \\
Self-collision weight $w_{\mathrm{self}}$ & $10$ \\
Contact-distance tolerance $d^\star$ & $2\,\mathrm{mm}$ \\
Joint-limit margin $\delta_q$ & $0.035\,\mathrm{rad}$ \\
Pre-/post-grasp gradient multipliers & $+0.5/-0.4$ \\
\bottomrule
\end{tabular}
\caption{\textbf{WBD grasp-generation parameters.} Optimization weights and tolerances.}
\label{tab:app-grasp-parameters}
\vspace{-4mm}
\endgroup
\end{table}

\subsection{Simulation Validation}

Because the optimizer minimizes a static surrogate objective, the WBD grasp
candidates are validated dynamically in two stages using mjlab~\cite{zakka2026mjlablightweightframeworkgpuaccelerated}. The torso is pinned to its
rest pose every substep, and the lower body is held fixed, so the torso frame is exactly
static; the object is a free body. Physics runs at
$1\,\mathrm{ms}$ with four substeps per control frame, an elliptic friction cone, and
surface friction of $\mu=0.6$ on the robot with priority over the object's.

Each simulation trial starts the arms at the pre-grasp, closes to the grasp candidate
configuration, presses to the post-grasp target, and holds there. Approach and press
last $40$ control frames each, followed by a brief $60$-frame settle and the
$300$-frame hold, with $4\,\mathrm{ms}$ per frame.
The gravity load and any perturbations are applied during the hold. A link is in contact when its
normal force is at least $0.1\,\mathrm{N}$.

\emph{Stage one} loads the object at $4\times$ weight and, over the first half of the
hold, sweeps a sustained $10\,\mathrm{N}$ force and $0.5\,\mathrm{N\,m}$ torque through
the six axis directions for $100\,\mathrm{ms}$ each, followed by an unperturbed holding phase. A
candidate passes if the object is still held in contact with at least two contact links at the end of the holding phase. 

\emph{Stage two} re-runs the survivors with no perturbations and normal gravity, and adds
pose gates, requiring the object to drift less than $0.10\,\mathrm{m}$ and rotate less
than $45^\circ$ over the hold. The configuration and object pose recorded at the end of
stage two are the settled values stored in the dataset and used at WBD-EX reset, so
training starts from poses settled under contact rather than raw optimizer solutions.

We refer throughout to a grasp's \emph{mode}, meaning
which arms must contact the object at the goal, as specified by its
optimization contact group. Mode $L$ requires left-arm contact, $R$ right-arm
contact, and $B$ both; single-arm goals also require the opposite arm to release
(\cref{app:wbdex-goals}).

\section{Voxelization and the WBD Grasp Dataset}
\label{app:voxel-cache}

\subsection{Dataset Contents}

Each object's accepted grasps are stored as a single dataset that WBD-EX loads at
startup. Per grasp, we keep the settled arm configuration $q_{\mathrm{sim}}$ (14
joints), the pressed post-grasp target $q_{\mathrm{post}}$ used as the servo setpoint
at reset, the settled object pose $T^{\mathrm{torso}}_{\mathrm{obj}}$ as a homogeneous
transform in the torso frame, and the grasp mode
$m\in\{L,R,B\}$.

\subsection{Position Voxelization}

We voxelize by the \emph{settled object center} expressed in the torso frame, taking
the translation of $T^{\mathrm{torso}}_{\mathrm{obj}}$ and binning it on a
$2\,\mathrm{cm}$ lattice anchored at the torso origin. A voxel's representative
position is the mean of its member grasp positions. Within
a voxel, grasps are grouped by mode, so a \emph{node} of the goal graph is an occupied
(voxel, mode) pair. Across the object set, this yields $1.0$--$5.5\times10^3$ occupied voxels
and $1.7$--$9.1\times10^3$ nodes; spheres sit at the low end and the elongated prisms at the
high end, because an elongated object reaches a wider spread of settled centers for the same
body region. The occupied region spans roughly $\pm 0.21\,\mathrm{m}$ laterally and
$0.03$--$0.45\,\mathrm{m}$ vertically.

\subsection{Goal Graph}

Mode compatibility and goal sampling follow \cref{sec:rl-training},
with $B\!\leftrightarrow\!B$ transitions also permitted. A mode
change within a single voxel is a legal zero-distance transition. 
Intermediate goals exclude the current (voxel, mode) node.

Far goals exclude the current voxel. If no legal-mode node lies within
the sampled distance, the sampler selects the nearest legal-mode node toward the far goal.
If no legal-mode node exists, the episode terminates as a dead end. A
curriculum ramps $d_{\max}$ linearly from $0.02\,\mathrm{m}$ to $0.08\,\mathrm{m}$ over the
first $5000$ iterations. Holds
earn no transition credit.

\subsection{Dataset Construction}
\label{app:cache-construction}

Candidate grasps enter the dataset only after the two-stage simulation filter of
\cref{app:grasp-gen}. We then apply a safety filter to the \emph{settled} pose with two
components. The settled object center must lie at least $9\,\mathrm{cm}$ above
the torso-frame origin, or $3\,\mathrm{cm}$ for objects whose mesh bounding-box height exceeds
$15\,\mathrm{cm}$. An exact on-body clearance test then drops any grasp whose object
surface sits farther than $5\,\mathrm{mm}$ from the torso.

\label{app:grasp-coverage}
Coverage measures sampled object-center positions around the torso.
For each object, we offset exterior torso-face centroids along their normals at six
evenly spaced distances between the object's minimum and maximum bounding-box half-extents.
We deduplicate positions on a $2\,\mathrm{cm}$ grid and keep those at torso-frame heights
$z\in[0.09,0.29]\,\mathrm{m}$. A reference position is covered if a grasp center
lies within the object's minimum half-extent. The per-object score is the fraction
covered, averaged equally across objects. For pooled coverage, we combine all grasp
centers and reference positions, deduplicate the latter on a $1\,\mathrm{cm}$ grid,
and use the $7.6\,\mathrm{cm}$ mean minimum half-extent as tolerance.

\section{WBD-EX Training Details}
\label{app:wbdex}

\subsection{Observations and Actions}
\label{app:wbdex-actions}

\Cref{tab:app-obs} lists the observation terms. The actor receives $89$ dimensions at the
current timestep, and the critic receives the same terms plus six privileged dimensions
giving the realized randomization of \cref{app:wbdex-dr}, for $95$ dimensions. Actor
observations are corrupted during training with additive uniform noise of
$\pm 0.005\,\mathrm{rad}$ on joint positions, $\pm 0.003\,\mathrm{m}$ on the goal offset
and object position, and $\pm 0.01$ on the rotation entries; the critic receives the same
terms uncorrupted.

\begin{table}[h]
\centering
\footnotesize
\setlength{\tabcolsep}{3pt}
\begin{tabular}{llr}
\toprule
Group & Term & Dim. \\
\midrule
\multirow{4}{*}{Proprioception}
 & Arm joint positions, minus default & 14 \\
 & Arm joint velocities & 14 \\
 & Servo error, commanded $-$ measured & 14 \\
 & Previous action & 14 \\
\midrule
\multirow{2}{*}{Command}
 & Hold flag, commanded and max.\ goal distance & 3 \\
 & Goal position relative to the object & 3 \\
\midrule
\multirow{2}{*}{Object}
 & Position and $6$-D rotation, torso frame & 9 \\
 & Linear and angular velocity, torso frame & 6 \\
\midrule
\multirow{4}{*}{Contact}
 & \shortstack[l]{Contacting-link fractions:\\left arm, right arm, torso} & 3 \\
 & Contacting links / total monitored links & 1 \\
 & Retention flag ($\rho_t$) & 1 \\
 & \shortstack[l]{Net contact-force magnitude\\(divided by $80\,\mathrm{N}$, capped at $2$)} & 1 \\
\midrule
Grasp mode & One-hot source and target mode & 6 \\
\midrule
\multicolumn{2}{l}{\textbf{Actor total}} & \textbf{89} \\
\midrule
\multirow{3}{*}{Privileged}
 & Object mass & 1 \\
 & Object center-of-mass offset & 3 \\
 & Mean sliding friction, per group & 2 \\
\midrule
\multicolumn{2}{l}{\textbf{Critic total}} & \textbf{95} \\
\bottomrule
\end{tabular}
\caption{\textbf{WBD-EX observation terms.} Current-timestep inputs, without history.
The actor uses the first five groups with training noise; the critic uses their
uncorrupted values plus the sampled dynamics parameters.}
\label{tab:app-obs}
\vspace{-2mm}
\end{table}

The action is a $14$-dimensional bounded increment on the arm joint targets. A raw
action is squashed through $\tanh$ and scaled by $0.025\,\mathrm{rad}$ per control
step, then accumulated into a servo setpoint that persists across steps. The
setpoint is clamped to lie within $0.35\,\mathrm{rad}$ of the measured position and
within the joint limits, and is linearly interpolated across the four physics
substeps. The lower body and waist are welded, so the torso frame is exactly static
and the arms are the only actuated degrees of freedom. Arm stiffness is
$k_p=20$; damping is derived per joint as $k_v \propto \sqrt{k_p}$.

\subsection{Rates, Episodes, Terminations}
\label{app:wbdex-terminations}

Physics runs at $200\,\mathrm{Hz}$ with a decimation of four, giving a
$50\,\mathrm{Hz}$ control rate. Training episodes last $40\,\mathrm{s}$
($2000$ control steps), and rewards are not scaled by the timestep. Contact is read
from a sensor between all links and the object.

An episode ends on time-out, which is bootstrapped, or on one of six failures: the
object falling below $0.5\,\mathrm{m}$ in the world frame; contact lost for longer than a grace window
of $24$ control steps ($0.48\,\mathrm{s}$), where contact is maintained if the arm contacts required by either the source or the target grasp mode are present, and torso contact alone does not count; the object leaving a
$0.55\,\mathrm{m}$ workspace radius or exceeding $6\,\mathrm{m/s}$ or
$40\,\mathrm{rad/s}$; the object coming to rest on the legs (disabled for large objects); a non-finite state; or a
dead end in the goal graph. The sparse failure penalty fires on the first five of
these; dead ends and time-outs are not penalized.

Each episode is initialized directly from the grasp dataset, with the arm configuration
and object pose written exactly from the sampled grasp and zero velocity. The servo target
is set instantly to that grasp's pressed target with no ramp. Grasps are
sampled as in \cref{sec:rl-training}. The only
continuous randomization at reset is a $\pm 5\,\mathrm{mm}$ jitter on movement goal
positions; hold goals are exact.

\subsection{Goal Completion Criteria}
\label{app:wbdex-goals}

We define the object as \emph{settled} when its linear and angular speeds are below
$0.12\,\mathrm{m/s}$ and $1.5\,\mathrm{rad/s}$, respectively.
A commanded goal is completed when all of the following hold at once: the object is within
$0.02\,\mathrm{m}$ of the commanded position, the object is settled,
the arms required by the goal grasp mode are in contact, and these conditions persist for $10$ control steps for movement goals or $25$ for
hold goals. A single-arm goal additionally requires the other arm to register no object
contact for five consecutive steps. Training and evaluation both apply this
requirement.

\subsection{Reward Details}
\label{app:wbdex-rewards}

\Cref{tab:app-reward-weights} lists the weights applied to the reward
terms defined in \cref{sec:rl-training} and below. Reward weights are fixed for
the whole of training; the two curricula modify the environment rather than
the objective.
For progress, both distance errors refer to the same commanded goal. Assigning a new
goal resets the distance baseline without generating progress.

\begingroup
The object's linear and angular speeds are denoted by $v_t$ and $\omega_t$.
Let $a_t$ be the raw policy action, $b_t=\tanh(a_t)$ the bounded action before scaling
to joint increments, $c_t$ the arm joint target, and $q_t$ the measured arm
joint positions. The five regularization penalties are:
\begin{itemize}[leftmargin=*, itemsep=2pt, topsep=3pt, parsep=0pt]
\item \emph{Unsupported motion}, $(1-\rho_t)\max(v_t - 5\,\mathrm{cm/s}, 0)^2$: penalizes linear object speeds above the threshold when $\rho_t=0$, discouraging unsupported motion.
\item \emph{Speed cap}, $\max(v_t - 0.35\,\mathrm{m/s}, 0)^2 + 0.02\max(\omega_t - 3\,\mathrm{rad/s}, 0)^2$: penalizes linear and angular speeds above their respective thresholds.
\item \emph{Action magnitude}, $\lVert b_t \rVert_2^2$: penalizes large bounded actions.
\item \emph{Action rate}, $\lVert b_t - b_{t-1} \rVert_2^2$: penalizes changes between consecutive bounded actions.
\item \emph{Servo lead}, $\sum_{j=1}^{14}[|c_{t,j}-q_{t,j}|-0.25\,\mathrm{rad}]_+^2$: penalizes joint targets more than $0.25\,\mathrm{rad}$ from the measured positions.
\end{itemize}
\endgroup

\begin{table}[h]
\centering
\footnotesize
\begingroup
\begin{tabular}{lr}
\toprule
Term & Coefficient \\
\midrule
Progress & $100$ \\
Goal-region arrival & $0.15$ \\
Transition completion & $20$ \\
Hold quality & $0.2$ \\
Arm neutrality & $0.5$ \\
Failure penalty & $-20$ \\
Unsupported motion & $-1$ \\
Speed cap & $-0.5$ \\
Action magnitude & $-10^{-4}$ \\
Action rate & $-10^{-3}$ \\
Servo lead & $-0.1$ \\
\bottomrule
\end{tabular}
\endgroup
\caption{\textbf{Reward coefficients.} Negative values denote penalties.}
\label{tab:app-reward-weights}
\vspace{-6mm}
\end{table}

\subsection{Domain Randomization}
\label{app:wbdex-dr}

Randomization ranges and schedules are listed in
\cref{tab:app-domain-rand}.

\begin{table}[h]
\centering
\footnotesize
\begingroup
\renewcommand{\arraystretch}{1.15}
\begin{tabular}{@{}>{\raggedright\arraybackslash}p{0.26\columnwidth}>{\raggedright\arraybackslash}p{0.68\columnwidth}@{}}
\toprule
Parameter & Range and sampling schedule \\
\midrule
Robot friction &
Per-geometry multiplier $\mathcal{U}(0.75,1.25)$, sampled on the first reset and fixed per environment. \\
\addlinespace[2pt]
Object mass and inertia &
Mass $0.4$--$1.2\,\mathrm{kg}$ for every object, with inertia co-scaled. Sampled at startup and fixed per environment. \\
\addlinespace[2pt]
Object center of mass &
Offset $\pm5\,\mathrm{mm}$ independently per axis. Sampled at startup and fixed per environment. \\
\addlinespace[2pt]
Perturbation force &
Magnitude $\mathcal{U}(0,F_{\max})$ and isotropic direction, applied at the object center. No applied torque. \\
\addlinespace[2pt]
Force duration &
$0.05$--$0.20\,\mathrm{s}$ per perturbation. \\
\addlinespace[2pt]
Cooldown &
$0.75$--$2.0\,\mathrm{s}$ between perturbations. \\
\addlinespace[2pt]
Force cap $F_{\max}$ &
$0$ for the first $2000$ iterations, then a linear ramp to $20\,\mathrm{N}$ over the next $8000$; $20\,\mathrm{N}$ thereafter. \\
\bottomrule
\end{tabular}
\endgroup
\caption{\textbf{WBD-EX domain randomization.} Dynamics parameters are fixed per environment, while perturbation forces ramp up during training.}
\label{tab:app-domain-rand}
\vspace{-4mm}
\end{table}

\subsection{Optimization}

Experts are trained in mjlab~\cite{zakka2026mjlablightweightframeworkgpuaccelerated} with PPO~\cite{schulman2017ppo} and the asymmetric actor--critic~\cite{pinto2018asymmetric}
inputs specified in \cref{app:wbdex-actions}. We use $64$ steps per environment, five epochs, and $16$
minibatches per update, discount $0.99$, GAE $\lambda = 0.95$, clip $0.2$, value-loss
coefficient $1$, gradient-norm clip $1$, and advantage normalization over the collected
batch. The learning rate starts at $10^{-3}$ under a KL-targeted adaptive schedule
with target $0.015$, clamped to $[10^{-5},10^{-2}]$; the second stage inherits the
pre-trained sphere policy's converged rate. The entropy coefficient is $0.005$ in the first stage and $0$ in
the second. The $\varnothing180$, $200$, and $220$\,mm sphere experts are not warm-started from another object; they are trained to $50{,}000$ iterations with the first-stage settings. Actor and critic are separate $512\times256$ ELU networks with empirical
observation normalization, and the policy is Gaussian with a learned standard deviation
initialized at $0.3$ and hard-clamped to $[0.05,0.6]$ in the first stage and $[0.05,0.4]$ in the second. Production runs use $8192$
environments on each of eight NVIDIA H100 GPUs, giving $65{,}536$ environments and roughly
$4.2\times10^{6}$ transitions per iteration.

\section{WBD-EX Regional Route Completion}
\label{app:route-eval}

WBD-EX policies are evaluated on fourteen standardized routes traversing six body regions (\cref{tab:app-routes}).

\subsection{Body Regions}

\begin{figure*}[t]
    \centering
    \includegraphics[width=\textwidth]{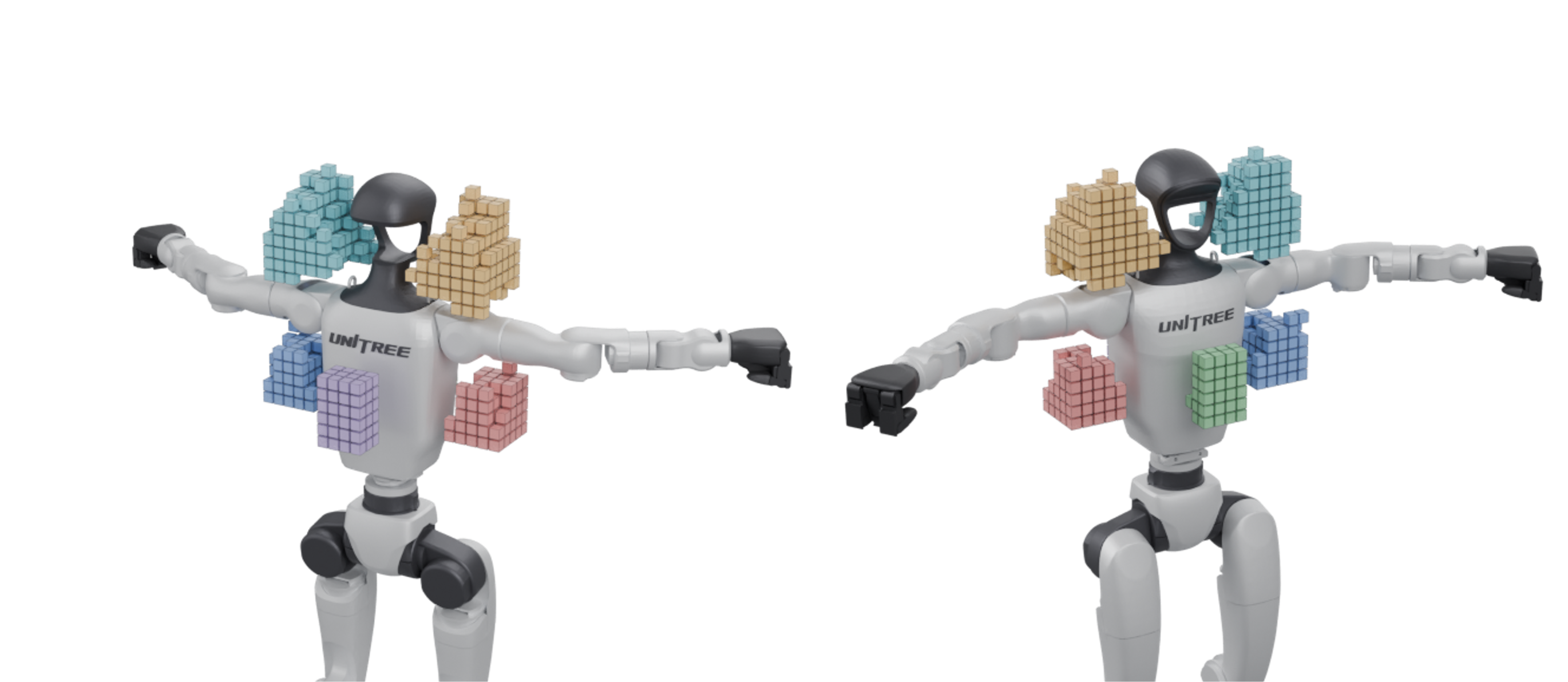}
    \caption{\textbf{Evaluation regions.}
    Six torso-frame goal regions shown from two viewpoints.
    Required grasp modes are
    both arms for front and back, and the corresponding single arm for the underarms and
    shoulders.}
    \label{fig:app-eval-regions}
    \vspace{-6mm}
\end{figure*}

Evaluation is defined over six named regions of the torso frame, applied to the settled
object center (\cref{fig:app-eval-regions}). The region bands are fixed across all
object sets so that scores are comparable; the voxels a given region admits differ
across objects only because the reachable grasp set does. Front and back occupy the three innermost $2\,\mathrm{cm}$ columns on
either side of the body within $\pm 3\,\mathrm{cm}$ laterally and
$0.090$--$0.19\,\mathrm{m}$ vertically, and both require a both-arm grasp. The underarms
lie beyond $\pm 0.11\,\mathrm{m}$ laterally, within
$-0.04$--$0.06\,\mathrm{m}$ front-to-back and $0.090$--$0.22\,\mathrm{m}$ vertically,
and require the corresponding single-arm grasp. The shoulders lie beyond
$\pm 0.05\,\mathrm{m}$ laterally, within $\pm 0.06\,\mathrm{m}$ front-to-back and
$0.29$--$0.49\,\mathrm{m}$ vertically, also requiring the corresponding single-arm grasp.
All expert evaluations use the completion and inactive-arm release
requirements in \cref{app:wbdex-goals}.

\subsection{Routes}

A route is a fixed sequence of regions. The environment is initialized inside the first
region, and each subsequent region becomes a commanded goal that must be reached in
order. We name a route by its
region sequence, abbreviating front as \texttt{F}, back as \texttt{B}, the left and
right underarms as \texttt{LA} and \texttt{RA}, and the left and right shoulders as
\texttt{LS} and \texttt{RS}. A route's name therefore states its itinerary: route
\texttt{F\_LA\_B} starts at the front, passes through the left underarm, and finishes at
the back. \Cref{tab:app-routes} lists all fourteen routes.

\begin{table}[h]
\centering
\footnotesize
\setlength{\tabcolsep}{1pt}
\begin{tabular}{ll}
\toprule
Route & Region sequence \\
\midrule
\multicolumn{2}{l}{\emph{Underarm}} \\
\texttt{F\_LA\_F} & front $\to$ left underarm $\to$ front \\
\texttt{F\_RA\_F} & front $\to$ right underarm $\to$ front \\
\texttt{LA\_F\_LA} & left underarm $\to$ front $\to$ left underarm \\
\texttt{RA\_F\_RA} & right underarm $\to$ front $\to$ right underarm \\
\addlinespace[2pt]
\multicolumn{2}{l}{\emph{Shoulder}} \\
\texttt{F\_LS\_F} & front $\to$ left shoulder $\to$ front \\
\texttt{F\_RS\_F} & front $\to$ right shoulder $\to$ front \\
\texttt{LS\_F\_LS} & left shoulder $\to$ front $\to$ left shoulder \\
\texttt{RS\_F\_RS} & right shoulder $\to$ front $\to$ right shoulder \\
\addlinespace[2pt]
\multicolumn{2}{l}{\emph{Back}} \\
\texttt{F\_LA\_B} & front $\to$ left underarm $\to$ back \\
\texttt{B\_RA\_F} & back $\to$ right underarm $\to$ front \\
\texttt{B\_LA\_B} & back $\to$ left underarm $\to$ back \\
\texttt{B\_RA\_B} & back $\to$ right underarm $\to$ back \\
\addlinespace[2pt]
\multicolumn{2}{l}{\emph{Loop}} \\
\texttt{F\_LA\_B\_RA\_F} & front $\to$ left underarm $\to$ back $\to$ right underarm $\to$ front \\
\texttt{B\_LA\_F\_RA\_B} & back $\to$ left underarm $\to$ front $\to$ right underarm $\to$ back \\
\bottomrule
\end{tabular}
\caption{\textbf{Evaluation routes.} Fourteen routes grouped by region. Each sequence includes the initial region;
subsequent regions are goals.}
\label{tab:app-routes}
\vspace{-2mm}
\end{table}

Regional route completion is the fraction of trials that reach every goal
of a prescribed route in order before failure or timeout; partial progress
along a route scores zero. The studies in \cref{app:expert-results,app:warmstart,app:culling}
specify their own checkpoint selection, route subsets, and initial-state filtering.

\begin{figure*}[t]
    \centering
    \includegraphics[width=\textwidth]{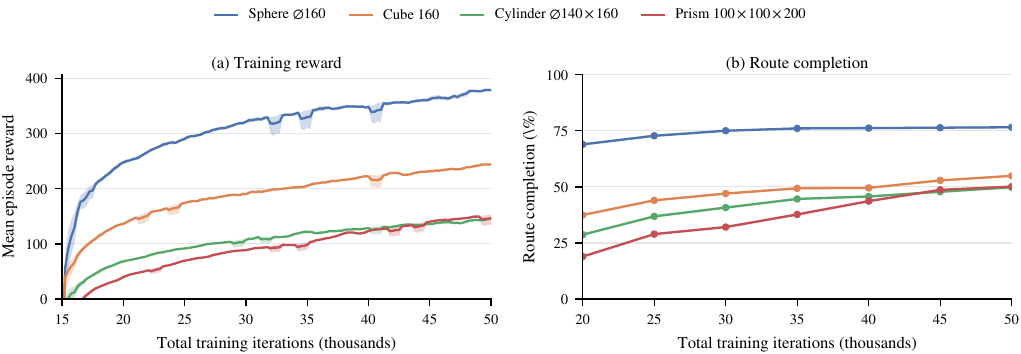}
    \vspace{-4mm}
    \caption{\textbf{Training progress for one expert per geometry class.} (a) Mean episode
    reward during training, averaged over up to $1{,}000$ logged updates;
    bands are approximate $95\%$ confidence intervals (CIs) from within-window block resampling.
    (b) Regional route completion, averaged equally over fourteen routes at
    $2{,}048$ trials per route; bands are pointwise $95\%$ evaluation CIs.
    One training seed per object.}
    \label{fig:app-expert-training}
    \vspace{-2mm}
\end{figure*}

\begin{figure*}[t]
    \centering
    \includegraphics[width=\textwidth]{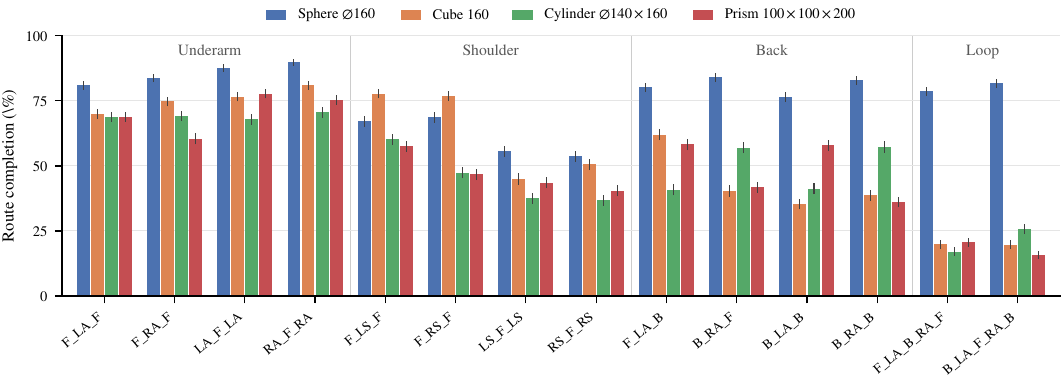}
    \vspace{-6mm}
    \caption{\textbf{Regional route completion.} Results at $50{,}000$ iterations for one
    expert per geometry class, grouped by the body region the route visits. Error bars are
    pointwise $95\%$ CIs from $2{,}048$ trials per route.
    One training seed per object.}
    \label{fig:app-expert-routes}
    \vspace{-6mm}
\end{figure*}

\subsection{Trials and Budget}

Every policy is scored against the same completion requirements regardless of how it was
trained, and the dwell, position threshold, and contact grace window match training.
Each route is evaluated with $2{,}048$ trials at a fixed seed and a $1{,}500$-step budget per
trial, with trials retired at the first termination, giving $28{,}672$ trials per
checkpoint over the fourteen routes. Evaluation retains the object-mass, friction, and
center-of-mass randomization of \cref{app:wbdex}, freezes the goal-distance curriculum at
its final value, and disables perturbations.

\subsection{Expert Evaluation Across Object Primitives}
\label{app:expert-results}
\label{app:expert-showcase}

One WBD-EX expert is trained per object and evaluated on all fourteen routes of
\cref{tab:app-routes} under the protocol of \cref{app:route-eval}.
\Cref{tab:app-expert-results} reports the measured fourteen-route mean for each object expert, grouped by the route groups
of \cref{tab:app-routes}, because objects do not rank from easy to hard along a single axis:
they fail in region-specific ways.

\begin{table}[!t]
\vspace{4mm}
\centering
\footnotesize
\setlength{\tabcolsep}{3pt}
\renewcommand{\arraystretch}{1.08}
\begin{tabular}{lrrrrr}
\toprule
& \multicolumn{4}{c}{Completion by region group (\%)} & All 14 \\
\cmidrule(lr){2-5}\cmidrule(lr){6-6}
Object (mm) & Underarm & Shoulder & Back & Loop & Mean (\%) \\
\midrule
Sphere $\varnothing$120 & 75.9 & 39.8 & 69.0 & 61.5 & 61.57 \\
Sphere $\varnothing$140 & 80.8 & 60.6 & 73.1 & 71.5 & 71.50 \\
Sphere $\varnothing$160 & 85.6 & 61.3 & 80.9 & 80.2 & 76.56 \\
Sphere $\varnothing$180 & 86.1 & 76.0 & 85.4 & 83.2 & 82.59 \\
Sphere $\varnothing$200 & 88.5 & 77.8 & 85.6 & 86.6 & 84.33 \\
Sphere $\varnothing$220 & 87.9 & 83.2 & 92.2 & 90.4 & 88.14 \\
\midrule
Cube 140 & 72.2 & 57.4 & 45.4 & 19.7 & 52.81 \\
Cube 160 & 75.6 & 62.6 & 44.1 & 19.8 & 54.91 \\
Cube 180 & 65.5 & 59.3 & 41.4 & 11.5 & 49.10 \\
Cube 200 & 61.3 & 69.7 & 12.5 & 0.0 & 40.98 \\
\midrule
Prism $100\!\times\!100\!\times\!200$ & 70.7 & 47.1 & 48.5 & 18.2 & 50.12 \\
Prism $80\!\times\!160\!\times\!220$ & 47.0 & 29.7 & 10.9 & 0.1 & 25.05 \\
Prism $140\!\times\!160\!\times\!240$ & 48.6 & 33.6 & 11.2 & 0.7 & 26.77 \\
\midrule
Cylinder $\varnothing120\!\times\!220$ & 66.9 & 33.2 & 29.2 & 6.2 & 37.82 \\
Cylinder $\varnothing140\!\times\!160$ & 69.2 & 45.5 & 49.1 & 21.3 & 49.83 \\
\bottomrule
\end{tabular}
\vspace{-1mm}
\caption{\textbf{WBD-EX expert results per object.} Equal-weight route means with
$2{,}048$ trials per route.}
\label{tab:app-expert-results}
\vspace{-10mm}
\end{table}

To characterize what a trained WBD-EX expert looks like, we take one object from each
geometry class: the $\varnothing160$\,mm sphere, the 160\,mm cube, the
$\varnothing140\times160$\,mm cylinder, and the $100\times100\times200$\,mm prism. We
report their training reward and regional route completion (\cref{fig:app-expert-training}) and their
final per-route profiles (\cref{fig:app-expert-routes}). All four are warm-started, so the
horizontal axes of \cref{fig:app-expert-training} start after the pre-trained sphere policy's $15{,}000$
iterations.

Completion is highest for spheres and lowest for the slender prisms and cylinder. The gap between geometry classes is largest on the most difficult trajectories: the four-goal loops, which chain the most transfers. Loop completion is near zero for the $200$\,mm cube and the two larger prisms ($80\times160\times220$\,mm and $140\times160\times240$\,mm; \cref{tab:app-expert-results}). Their large size relative to the G1's limbs contributes to this difficulty. Although route completion differs across object shapes and sizes, all experts supply trajectories for distillation. Continued improvement of expert success across all object shapes and sizes is left to future work.

\begin{table*}[t]
\centering
\footnotesize
\setlength{\tabcolsep}{4pt}
\begin{tabular}{l r >{\columncolor{black!7}}r r >{\columncolor{black!7}}r r >{\columncolor{black!7}}r r >{\columncolor{black!7}}r r >{\columncolor{black!7}}r}
\toprule
& \multicolumn{2}{c}{Underarm} & \multicolumn{2}{c}{Shoulder} & \multicolumn{2}{c}{Back}
& \multicolumn{2}{c}{Loop} & \multicolumn{2}{c}{All 14 routes} \\
\cmidrule(lr){2-3}\cmidrule(lr){4-5}\cmidrule(lr){6-7}\cmidrule(lr){8-9}\cmidrule(lr){10-11}
Object (mm) & Scr. & Warm & Scr. & Warm & Scr. & Warm & Scr. & Warm & Scr. & Warm \\
\midrule
Prism $100\!\times\!100\!\times\!200$ & 35.3 & \textbf{70.7} & 27.0 & \textbf{47.1} & 18.9 & \textbf{48.5} & 0.8 & \textbf{18.2} & 23.30 & \textbf{50.12} \\
Cube 140 & 67.5 & \textbf{72.2} & 55.2 & \textbf{57.4} & 46.2 & 45.4 & 15.6 & \textbf{19.7} & 50.48 & \textbf{52.81} \\
Cube 200 & 55.1 & \textbf{61.3} & 68.9 & 69.7 & 0.3 & \textbf{12.5} & 0.0 & 0.0 & 35.53 & \textbf{40.98} \\
Cylinder $\varnothing140\!\times\!160$ & 68.2 & 69.2 & 29.6 & \textbf{45.5} & \textbf{56.7} & 49.1 & \textbf{25.2} & 21.3 & 47.74 & \textbf{49.83} \\
Sphere $\varnothing$140 & 81.1 & 80.8 & 39.1 & \textbf{60.6} & 72.7 & 73.1 & 71.8 & 71.5 & 65.37 & \textbf{71.50} \\
Sphere $\varnothing$160 & \textbf{89.4} & 85.6 & 42.1 & \textbf{61.3} & \textbf{83.8} & 80.9 & \textbf{83.9} & 80.2 & 73.51 & \textbf{76.56} \\
\bottomrule
\end{tabular}
\caption{\textbf{Warm-start versus from-scratch experts by body region}, at 50{,}000 matched
total iterations with $2{,}048$ trials per route. Entries are mean regional
route completion (\%), equally weighted over each group's routes. Bold marks the higher score when
the $95\%$ evaluation CI for the difference excludes zero. One training seed per condition.}
\label{tab:app-warmstart}
\vspace{-4mm}
\end{table*}

\section{Warm-Start Versus Training from Scratch}
\label{app:warmstart}

\Cref{sec:rl-training} describes the two-stage procedure: each warm-started target expert
is initialized from a pre-trained sphere expert's complete training state. Here, we
compare it against training the same object from scratch.

\subsection{Setup and Matched Budgets}
\label{app:warmstart-setup}

We compare warm-started and from-scratch experts for six objects: a $100\times100\times200$\,mm prism, cubes of side 140 and 200\,mm, a
$\varnothing140\times160$\,mm cylinder, and spheres of diameter 140 and 160\,mm.
We warm-start these experts from policies pre-trained on spheres
with diameters of $200$--$220$\,mm. All share one training recipe and the
randomization of \cref{app:wbdex-dr}, and both conditions switch to the second-stage settings of \cref{app:wbdex} at 15{,}000 total iterations.
Budgets are matched at 50{,}000 total iterations. The warm-started expert is charged its pre-trained sphere policy's 15{,}000
iterations plus 35{,}000 on its own object, so it reaches the comparison point having seen
its target object for 15{,}000 fewer iterations. All six are
compared on all fourteen routes.

\subsection{Results by Body Region}
\label{app:warmstart-results}

\begin{figure*}[t]
    \centering
    \includegraphics[width=0.96\textwidth]{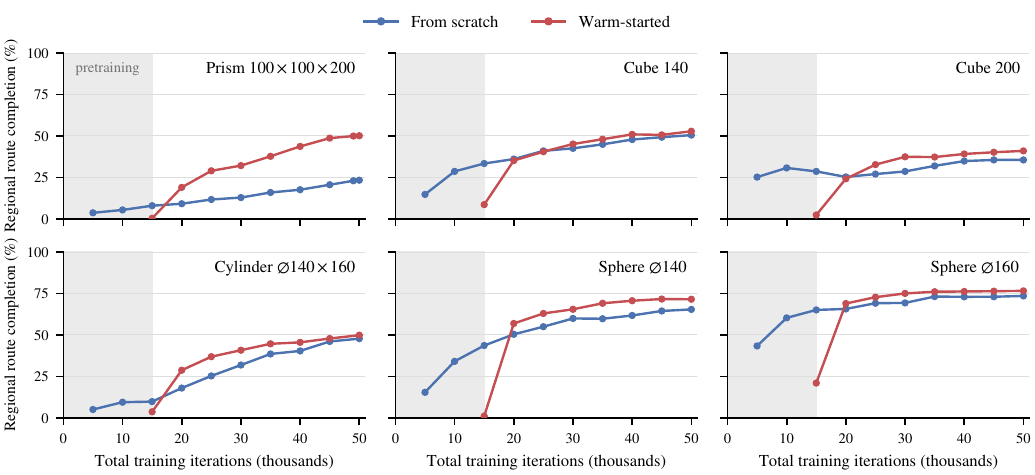}
    \caption{\textbf{Warm-started and from-scratch learning curves.}
    Warm-started curves include $15{,}000$ sphere pre-training
    iterations (shaded); their first point precedes target-object training.
    Markers show mean regional route completion, weighting all fourteen routes equally.
    Bars show pointwise $95\%$ evaluation CIs (binomial trial uncertainty
    or evaluation-batch bootstrap), often smaller than the markers. One training seed per condition.}
    \label{fig:app-warmstart-history}
    \vspace{-6mm}
\end{figure*}

\Cref{tab:app-warmstart} breaks the comparison down by body region, and
\cref{fig:app-warmstart-history} shows the corresponding training progress.
All six warm-started experts have higher mean regional route completion, although some regional
scores are lower.
Shoulder completion is higher for every warm-started expert.
The prism has the largest aggregate difference,
from $23.30\%$ to $50.12\%$, including loop completion of $0.8\%$ versus $18.2\%$.
For cube 200, mean completion across the four back routes is $0.3\%$ from scratch
and $12.5\%$ with warm-starting. Both conditions have $0.0\%$ loop completion.
Other regions show smaller differences or regressions. The warm-started
cylinder has lower back and loop completion, and the $\varnothing160$\,mm sphere has
lower underarm, back, and loop completion despite its higher overall mean.

\subsection{Training Cost}
\label{app:warmstart-cost}

The $15{,}000$-iteration sphere pre-training cost is shared by all experts
initialized from that policy. For $N$ target experts, each post-trained for $35{,}000$
iterations, the average budget is $35{,}000 + 15{,}000/N$ iterations per expert,
compared with $50{,}000$ iterations per expert trained from scratch.
Counting the six objects here with their two pre-trained sphere policies,
the total is 240{,}000 iterations against 300{,}000, a $20\%$ reduction
in aggregate iterations with all six final means higher.

Representative runs on
eight NVIDIA H100 GPUs use $15{,}000$ sphere pre-training iterations at approximately $270$
H100 GPU-hours, followed by $35{,}000$ post-training iterations per object at
$660$--$815$ H100 GPU-hours. Training an expert from scratch for $50{,}000$ iterations
costs approximately $1{,}030$ H100 GPU-hours.
The shared pre-training contribution is therefore approximately
$270/N$ H100 GPU-hours per target. In these runs, the sphere also has the lowest
cost per iteration, complementing its shorter pre-training budget.
The sphere's compact contact patches produce fewer simultaneous contact constraints, reducing the simulator's contact-solver cost.
These are GPU costs for the runs that produced these experts.

\Cref{fig:app-warmstart-history} shows several warm-started experts crossing the
from-scratch endpoint well before their own budget is spent (the prism after $10{,}000$
target iterations and both spheres within $15{,}000$),
suggesting additional savings if post-training stops at matched regional route completion.

The initialization transfers the actor, critic, optimizer state, observation normalizers, and
curriculum progress together.

\section{Escapability-Based Selection of Expert Initializations}
\label{app:culling}

Grasp simulation validation (\cref{app:grasp-gen}) certifies that a grasp stably holds the object, not that a trained expert can reposition the object from it, and experts can transition out of some validated grasps far more readily than others. Down-weighting low-quality grasps during WBD-EX training underperformed training on the full grasp set, so we instead score each grasp with its trained expert and filter after training, when selecting initial states for trajectory collection (\cref{app:distill-data}). Within each voxel, we remove the lowest-scoring grasps while preserving occupied voxel--grasp-mode pairs.

Policy weights, the goal pool, and intermediate routing remain unchanged. We report
regional route completion to characterize the effect of this
selection using the same completion requirements as the expert evaluations.

\subsection{Escapability Scoring}
\label{app:culling-score}

For a grasp $g$, \emph{escapability} is the fraction of transitions attempted from $g$, each to a single
commanded goal, that reach the commanded position and grasp mode under the requirements of
\cref{app:route-eval}. Targets are drawn over the eligible mode-compatible
position voxels $6$--$8\,\mathrm{cm}$ from the source voxel's representative
position, with a budget of $3\,\mathrm{s}$ ($150$ control steps) per attempt.
The score is specific to the frozen WBD-EX expert. We evaluated all $2{,}510{,}864$ grasps across the fifteen objects with $128$
sampled targets per grasp, for $3.2\times10^{8}$ single-goal transition trials.

\begin{figure*}[t]
    \centering
    \includegraphics[width=\textwidth]{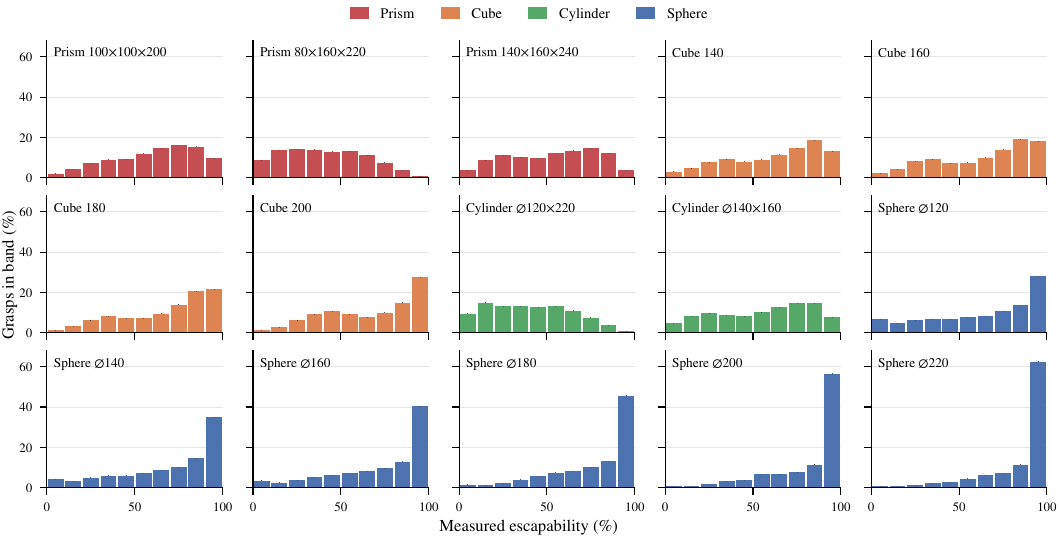}
    \caption{\textbf{Grasp escapability distributions.} Bars give the
    percentage of that object's grasps falling in each ten-point escapability band, colored by
    geometry class. Error bars are pointwise $95\%$ bootstrap CIs over grasps,
    with measured scores held fixed; counts are exact for the scored pools.}
    \label{fig:app-culling-esc}
    \vspace{-2mm}
\end{figure*}

\Cref{fig:app-culling-esc} shows that the distribution's shape, not just its mean, varies
strongly with geometry. The two most slender objects
(the $\varnothing120\times220$\,mm cylinder and the $80\times160\times220$\,mm prism)
have nearly flat distributions with mean escapability near $40\%$ and only about $1\%$ of
their grasps in the top band. Sphere grasps are increasingly concentrated at high
escapability with diameter, from $28\%$ of grasps in the top band at $\varnothing120$ to
$62\%$ at $\varnothing220$, with means of $64\%$ and $86\%$. Thus, grasps that pass
grasp simulation filtering can still have low transition success under their fully trained expert, and the distribution
of this success varies across the object-specific grasp datasets.

\subsection{Removal Threshold Selection}
\label{app:culling-select}

\begin{figure*}[t]
    \centering
    \includegraphics[width=\textwidth]{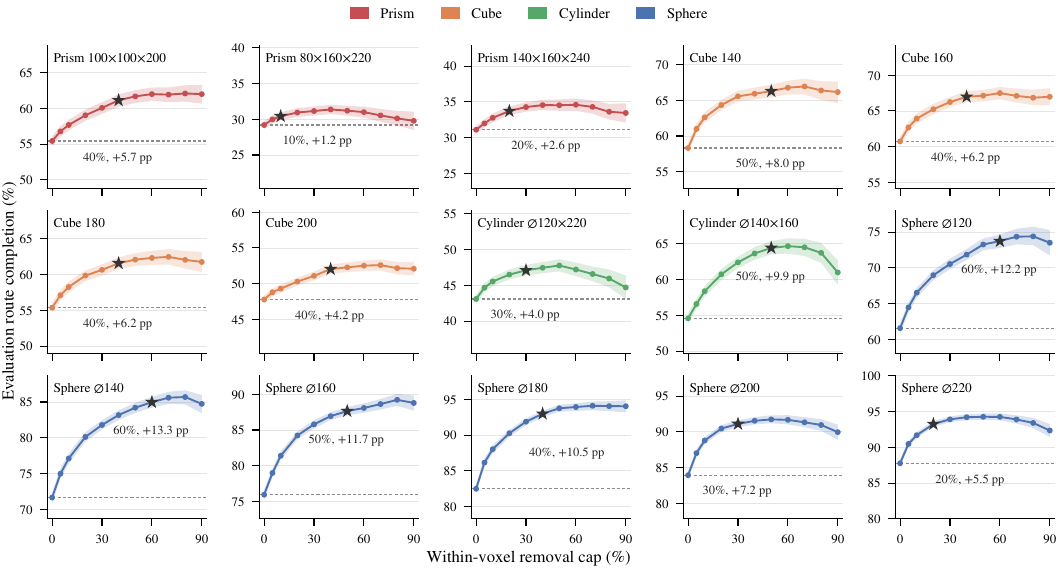}
    \caption{\textbf{Regional route completion after grasp filtering.}
    Mean completion over twelve equally weighted routes, excluding loops.
    Dashed lines show unfiltered baselines; stars mark selected caps and gains in percentage points.
    Bands are pointwise $95\%$ paired bootstrap CIs over evaluation trials.
    Vertical axes span twenty percentage points.}
    \label{fig:app-culling-sweep}
    \vspace{-6mm}
\end{figure*}

Within each original $2\,\mathrm{cm}$ position voxel, we remove the lowest-scoring grasps up
to a nominal percentage cap, keeping at least twenty grasps where available and preserving every original
voxel--grasp-mode pair. 
We use the twelve non-loop evaluation routes of
\cref{tab:app-routes}.
We re-score each expert on the saved evaluation trials whose
starts survive the cut, sweeping the cap from $0$ to $90\%$.
\Cref{fig:app-culling-sweep} shows mean regional route completion
for each object across removal caps.
We use $2{,}048$ bootstrap resamples of evaluation trials, shared across routes and removal caps.

At the adopted cuts, the per-object means in \cref{fig:app-culling-sweep}
improve by $1.2$--$13.3$ percentage points over the full grasp dataset baseline
across all fifteen objects. The cut preserves every occupied
voxel--grasp-mode pair and preserves at least $98.8\%$ of each object's average per-joint span
at every cap up to $70\%$. It preserves spatial and grasp-mode coverage while changing the
distribution of grasp configurations within each voxel.

The removal each object tolerates tracks the shape of its escapability distribution. Objects
with flat distributions take small cuts for small gains: $10\%$ removal for $1.2$ points on the
slender prism and $30\%$ for $4.0$ points on the slender cylinder, against $20$--$60\%$ removal
for $5$ to $13$ points on the spheres. Where escapability is uniformly moderate, there is little
low-escapability tail to remove, and the curve for the slender prism is nearly flat across the
whole sweep. At larger removal caps, completion for the slender cylinder and the two largest
spheres peaks and then declines.

We adopt, per object, the smallest cap within one percentage point of the best mean available
to it, subject to a $70\%$ ceiling and to preserving at least $95\%$ of the original average
joint-span coverage, and we keep the full grasp dataset where the interval on the gain is not
positive. 
The adopted caps range from $10$ to $60\%$. 
The gains above are measured on the trials used to select the caps.
The separate trajectory collection comparison below measures usable data yield directly.

\subsection{Trajectory Collection Yield}
\label{app:culling-efficiency}

We compare trajectory collection from the full and escapability-filtered
grasp datasets using $13{,}440$ collection attempts per condition across all fifteen objects.
Both conditions use the same frozen experts, set of eligible goals, sampling mixture,
randomization, and quality filter (\cref{app:distill-data}). Each trajectory follows successive
voxel-to-voxel goals until failure or a $120$\,s timeout. A trajectory is usable if, after removing the termination tail, it still passes the quality filter.
Each condition splits its attempts between the two route samplers of \cref{app:distill-data}. Combining both samplers, usable yield increases from approximately
$29.0\%$ to $32.5\%$ ($+3.5$ percentage points), and usable frames per attempt from
approximately $938$ to $1{,}053$ ($+12\%$).

The fraction of attempts failing within the first $10$\,s falls from
approximately $26.7\%$ to $22.9\%$. Accepted trajectories contain approximately $3{,}200$ frames
under both conditions, suggesting that filtering mainly increases the number of accepted trajectories. The improvement in collection yield is consistent but modest. We use the filtered grasp datasets for trajectory
collection because they raise the usable data per collection attempt without notably reducing the
coverage of the grasp datasets.

\section{WBD Controller Distillation Details}
\label{app:distill}

\Cref{sec:refinement} gives the WBD controller's architecture and its guidance mechanism.
Here we specify the network settings, training data, sampling
recipe, and inference procedure.

\subsection{Architecture and Optimization}
\label{app:distill-optimization}

We follow the diffusion-and-inverse-dynamics formulation and published
state-encoder/U-Net configuration of~\cite{yin2025dexteritygenfoundationcontrollerunprecedented},
adapting the representation and conditioning to whole-body manipulation.
Our U-Net uses one middle block, two linear layers per block, and a
$256$-dimensional timestep embedding.

Each arm contributes four keypoints: one at the elbow and three on
the hand, at the thumb, index finger, and middle finger.
Forward kinematics expresses these points in the torso frame using the fixed
closed-fist configuration.

Each history entry contains $24$ torso-frame arm-keypoint coordinates,
$14$ measured joint positions, $14$ applied joint targets, and $14$ joint tracking
errors. Four entries at offsets $(-15,-10,-5,0)$, in $50$\,Hz control steps,
and five static geometry features form the $269$-dimensional diffusion input;
no task-mode embedding is used. The model predicts $48$ keypoint-displacement
coordinates at future offsets $(5,10)$, with targets defined as future measured
keypoint positions minus their current measured positions.
The inverse model takes the current history entry, predicted motions, and geometry
features ($119$ inputs), using width $1536$ and four two-layer residual blocks
to output $14$ joint-target increments.

We train both networks jointly for $130{,}000$
AdamW updates with a global batch of $2{,}560$ training samples and base learning rate
$3\times10^{-4}$, including a final $10{,}000$-update cosine cooldown. The inverse
model is supervised on an equal mixture of recorded expert motions and
diffusion-generated motions, using the expert's realized joint-target increment at the same state for both: the applied target after action scaling, tracking-error clamping, and joint limits, minus the current target.
Diffusion training uses $100$ timesteps with a cosine schedule
(offset $0.008$, $\beta$ capped at $0.999$).

Training the simulation and real-world controllers uses approximately $22$ H100
GPU-hours per model across eight H100 GPUs.

\subsection{Trajectory Collection}
\label{app:distill-data}

Each frozen WBD-EX expert starts from its object's escapability-filtered
grasp dataset (\cref{app:culling}) and follows successive position-and-grasp-mode goals
without resetting between goals. Trajectories are recorded at $50\,\mathrm{Hz}$ until failure or
a $120\,\mathrm{s}$ timeout.

\emph{Random voxel-sampled routes} sample eligible goal voxels and compatible
grasp modes, with initial left/right/both modes in a $1{:}1{:}2$ ratio.
\emph{Random region-graph routes} sample initial and target grasps from voxels in connected
body regions. The first transfer follows one of sixteen directed region pairs:
front--shoulder, front--underarm, shoulder--underarm, and underarm--back, in both directions
and on both sides; subsequent goals follow the voxel goal graph (\cref{app:voxel-cache}).
Voxel-sampled routes with at least four completed goals also supply a longer-trajectory subset.
Both samplers balance the $2$--$4$, $4$--$6$, and $6$--$8\,\mathrm{cm}$ goal-distance bands,
while keeping the maximum distance used to build the goal graph at $8\,\mathrm{cm}$.

We keep the usable portions of failed trajectories, discarding the final $100$
frames before failure termination. Clean timeouts are left untrimmed, and we require at least two completed
goals and $50$ frames. Stalled intervals are removed using
$50$-frame windows in which the goal is unchanged, all joint spans are within $0.02\,\mathrm{rad}$, the object
position bounding box is within $5\,\mathrm{mm}$, and the maximum pairwise object rotation
is within $0.1\,\mathrm{rad}$.
Each training sample is drawn from a contiguous $26$-frame
trajectory segment that supplies the required observation history and future supervision.
This segment never crosses a removed interval or an episode boundary.

\subsection{Sampling and Augmentation}
\label{app:distill-sampling}

Batches mix training samples from random voxel-sampled routes,
region-graph routes, and the longer-trajectory subset. Each $320$-sample per-GPU batch draws $125$, $165$, and $30$ samples from these three sources and approximately equal numbers of samples from the $2$--$4$, $4$--$6$, and $6$--$8$\,cm goal-distance bands, with objects equally represented, while spatial and grasp-mode sampling promote coverage of
body regions and single-arm grasp goals. Training samples are drawn from eligible timesteps
throughout each trajectory; spatial sampling uses the object's position at
each sample's current frame.
We split trajectories and their derived samples by initial grasp using a fixed hash,
assigning approximately $5\%$ of grasp groups to validation.

Half of all training samples are reflected across the sagittal plane, applied as one
coherent transform over history, actions, and future targets in physical coordinates before
normalization, with validation and evaluation left unaugmented.
All trajectories are collected under randomized dynamics.
Mass, center-of-mass, and friction randomization follow
\cref{app:wbdex-dr}; the experts' perturbations remain enabled at their trained
$20\,\mathrm{N}$ cap. The randomized parameters are not controller inputs.

\subsection{Simulation Evaluation Routes}
\label{app:sim-region-walk}

For the simulation comparisons in \cref{fig:sim_results}, we reuse the
WBD-EX evaluation regions and goal-completion criteria (\cref{app:route-eval}). Each trial follows a continuous sequence over the region connections of \cref{app:distill-data}, sampling the next adjacent region uniformly after each completed transfer. No external perturbation forces are applied.

\begingroup
\subsection{Command-Smoothing Comparison}
\label{app:command-smoothing}

We test whether simple temporal smoothing can reproduce WBD's gains under
command corruption.
We compare the noisy expert, exponential moving average (EMA) smoothing of its
commands, and WBD refinement. For corrupted raw arm commands $u_t$, EMA applies
$z_t=(1-\beta)z_{t-1}+\beta u_t$ before action bounding and scaling, initialized
with $z_0=u_0$ and retained across goal changes.
We select $\beta=0.5$ from $\{0.05,0.1,0.25,0.5\}$ by mean completed regional
transfers on $64$ separate validation starts per training object and noise level,
weighting the $13$ objects and two noise levels equally. Test starts and their
mirrored families are excluded from selection.
The selected coefficient, $\beta=0.5$, also gives the highest
validation goal count among the tested EMA settings.

\begin{table}[h]
\centering
\footnotesize
\setlength{\tabcolsep}{3pt}
\renewcommand{\arraystretch}{1.1}
\begin{tabular}{clcc}
\toprule
$\alpha$ & Method & Goals / trial & Time to failure (s) \\
\midrule
$2.5$ & Noisy expert & $8.47$ & $29.1$ \\
 & Noisy + EMA & $7.82$ & $20.5$ \\
 & Noisy + WBD & $\mathbf{23.79}$ & $\mathbf{86.1}$ \\
\midrule
$4$ & Noisy expert & $6.24$ & $24.2$ \\
 & Noisy + EMA & $6.32$ & $17.6$ \\
 & Noisy + WBD & $\mathbf{21.79}$ & $\mathbf{64.6}$ \\
\bottomrule
\end{tabular}
\caption{\textbf{Command smoothing versus WBD refinement.}
Means weight the $13$ training objects equally, and WBD values are averaged over three training seeds.}
\label{tab:app-command-smoothing}
\vspace{-3mm}
\end{table}

This separate comparison repeats front $\rightarrow$ left underarm $\rightarrow$ front $\rightarrow$ right underarm at $\alpha\in\{2.5,4\}$, using $256$
matched starts per object and terminating at the first failure or $600$\,s.
We use the goals-per-trial metric defined in \cref{sec:sim-experiments}.
All failures remain included.
WBD uses guidance $100$; the expert checkpoints, action scaling, joint limits,
and task physics are shared across conditions.

The selected EMA filter yields similar goal counts and shorter time to
failure than the noisy expert. WBD improves both metrics over this filter at both
noise levels (\cref{tab:app-command-smoothing}).
\endgroup

\subsection{Real-Time Inference}
\label{app:distill-inference}

Each inference update samples $16$ candidate future motions using $12$ DDIM denoising steps.
Each candidate predicts keypoint displacements $100$ and $200$\,ms into the future
from the robot's current measured configuration. Guidance specifies displacement
from the current commanded configuration to the requested configuration.
These displacements are compared directly; the model also receives the joint
tracking error between commanded and measured configurations.

\textit{Command chain.} Let $c$ be the current applied joint target and $q$ the measured joint configuration. In simulation, the noisy raw expert action $\tilde a=a+\epsilon$, $\epsilon\sim\mathcal U(-\alpha,\alpha)^{14}$, defines the requested configuration $q^{\mathrm{req}}=\Pi(c+6\cdot0.025\tanh\tilde a)$, where $\Pi$ applies the $0.35$\,rad tracking-error clamp around $q$ and the joint limits. On hardware, $q^{\mathrm{req}}$ is the retargeted operator configuration clipped to the joint limits. The guidance displacement is $\mathbf d=\mathrm{FK}(q^{\mathrm{req}})-\mathrm{FK}(c)$, with no additional magnitude cap on this input displacement on hardware. The inverse model predicts an increment, giving $\hat c=c+\Delta\hat c$. In simulation, $\hat c$ is mapped back to a raw action $\hat a=\operatorname{atanh}(\mathrm{clip}((\hat c-c)/0.025,\pm0.999))$, clipped to $[\tilde a-2.5,\tilde a+2.5]$, and applied through the experts' action pipeline $\Pi(c+0.025\tanh\hat a)$; this $\pm2.5$ raw-action radius is the bounded correction in \cref{sec:sim-experiments}. On hardware, the predicted output increment is clipped to $\pm0.025$\,rad per joint; the publisher then applies its configured joint-speed limit, a $0.35$\,rad tracking-error limit, and the joint limits.

With $\mathbf d\in\mathbb{R}^{24}$ the guidance displacement defined above,
for predicted keypoint displacements $\Delta\mathbf{x}_5$ and $\Delta\mathbf{x}_{10}$
at the two future offsets, guidance minimizes
\begin{equation*}
D=\|\Delta\mathbf{x}_5-\mathbf d/2\|_2+
                    \|\Delta\mathbf{x}_{10}-\mathbf d\|_2.
\end{equation*}
Each norm spans all keypoint coordinates in physical space. At each denoising step, the guidance weight scales a gradient step that reduces $D$.
We select the generated candidate with the smallest summed squared error to these targets.
The inverse-dynamics model then predicts a joint-target increment from the selected
keypoint motion and the most recent state in the same observation history used for diffusion.

By the time this increment reaches the $50$\,Hz command publisher, the applied target
may have changed. Let $c^{\mathrm{obs}}$ be the applied target in that observation
and $\hat c$ the corresponding predicted target. The publisher adds the already-computed
increment $\hat c-c^{\mathrm{obs}}$ to its latest applied target $c^{\mathrm{pub}}$:
\begin{equation}
    c^{\mathrm{new}}
    = c^{\mathrm{pub}} + (\hat c-c^{\mathrm{obs}}).
\end{equation}
Between predictions, the publisher continues toward the resulting target without
applying the same increment again. We construct observation histories from timestamped
measurements and applied commands at the offsets in \cref{app:distill-optimization},
independently of inference timing.
On hardware, inference runs on an NVIDIA RTX 5090 GPU on the operator workstation and is capped at $50$\,Hz. With these sampler settings, one inference update took $11.8$\,ms median ($17.3$\,ms $95$th percentile), updates arrived at $34.9$\,Hz on average, and the delay from state publication to the robot's command buffer was $50.1$\,ms median ($73.7$\,ms $95$th percentile), excluding sensing and motor latency.

%% file: references.bib
@article{viral2025,
  author  = {Tairan He and Zi Wang and Haoru Xue and Qingwei Ben and Zhengyi Luo and Wenli Xiao and Ye Yuan and Xingye Da and Fernando Casta{\~n}eda and Shankar Sastry and Changliu Liu and Guanya Shi and Linxi Fan and Yuke Zhu},
  title   = {{VIRAL}: Visual Sim-to-Real at Scale for Humanoid Loco-Manipulation},
  journal = {arXiv preprint arXiv:2511.15200},
  year    = {2025}
}

@article{sonic2025,
  author  = {Zhengyi Luo and Ye Yuan and Tingwu Wang and Chenran Li and Fernando Casta{\~n}eda and Sirui Chen and Zi-Ang Cao and Jiefeng Li and David Minor and Qingwei Ben and Jinhyung Park and David Sami and Zi Wang and Xingye Da and Runyu Ding and Cyrus Hogg and Lina Song and Edy Lim and Eugene Jeong and Tairan He and Haoru Xue and Wenli Xiao and Simon Yuen and Jan Kautz and Yan Chang and Umar Iqbal and Linxi Fan and Yuke Zhu},
  title   = {{SONIC}: Supersizing Motion Tracking for Natural Humanoid Whole-Body Control},
  journal = {Science Robotics},
  volume  = {11},
  number  = {117},
  pages   = {eaed4592},
  year    = {2026},
  doi     = {10.1126/scirobotics.aed4592}
}

@article{sumo2026,
  author  = {John Z. Zhang and Maks Sorokin and Jan Br{\"u}digam and Brandon Hung and Stephen Phillips and Dmitry Yershov and Farzad Niroui and Tong Zhao and Leonor Fermoselle and Xinghao Zhu and Chao Cao and Duy Ta and Tao Pang and Jiuguang Wang and Preston Culbertson and Zachary Manchester and Simon {Le Cl{\'e}ac'h}},
  title   = {{Sumo}: Dynamic and Generalizable Whole-Body Loco-Manipulation},
  journal = {arXiv preprint arXiv:2604.08508},
  year    = {2026}
}

@inproceedings{psi02026,
  author    = {Songlin Wei and Hongyi Jing and Boqian Li and Zhenyu Zhao and Jiageng Mao and Zhenhao Ni and Sicheng He and Sheng Zang and Xiawei Liu and Kaidi Kang and Jie Liu and Weiduo Yuan and Marco Pavone and Di Huang and Yue Wang},
  title     = {{$\Psi_0$}: An Open Foundation Model Towards Universal Humanoid Loco-Manipulation},
  booktitle = {Robotics: Science and Systems ({RSS})},
  year      = {2026},
  doi       = {10.15607/RSS.2026.XXII.021}
}

@inproceedings{yin2025dexteritygenfoundationcontrollerunprecedented,
      title={{DexterityGen}: Foundation Controller for Unprecedented Dexterity}, 
      author={Zhao-Heng Yin and Changhao Wang and Luis Pineda and Francois Robert Hogan and Chaithanya Krishna Bodduluri and Akash Sharma and Patrick Lancaster and Ishita Prasad and Mrinal Kalakrishnan and Jitendra Malik and Mike Lambeta and Tingfan Wu and Pieter Abbeel and Mustafa Mukadam},
      year={2025},
      eprint={2502.04307},
      archivePrefix={arXiv},
      primaryClass={cs.RO},
  booktitle={Robotics: Science and Systems (RSS)},
  doi={10.15607/RSS.2025.XXI.103}
}

@article{kedia2026simtoolrealobjectcentricpolicyzeroshot,
      title={{SimToolReal}: An Object-Centric Policy for Zero-Shot Dexterous Tool Manipulation}, 
      author={Kushal Kedia and Tyler Ga Wei Lum and Jeannette Bohg and C. Karen Liu},
      year={2026},
      eprint={2602.16863},
      archivePrefix={arXiv},
      primaryClass={cs.RO},
  journal={arXiv preprint arXiv:2602.16863}
}

@article{lee2026adeptacceleratingdexteritypretraining,
      title={{ADEPT}: Accelerating Dexterity via Pre-Training and Post-Training using Reinforcement Learning}, 
      author={Jayjun Lee and Jessica Yin and Asif Rana and Nicholas Blauch and Sam Mady and Mohak Bhardwaj and Nima Fazeli and Nathan Ratliff and Karl Van Wyk and Ankur Handa},
      year={2026},
      eprint={2608.19182},
      archivePrefix={arXiv},
      primaryClass={cs.RO},
  journal={arXiv preprint arXiv:2608.19182}
}

@inproceedings{ye2025dex1blearning1bdemonstrations,
      title={{Dex1B}: Learning with {1B} Demonstrations for Dexterous Manipulation}, 
      author={Jianglong Ye and Keyi Wang and Chengjing Yuan and Ruihan Yang and Yiquan Li and Jiyue Zhu and Yuzhe Qin and Xueyan Zou and Xiaolong Wang},
      year={2025},
      eprint={2506.17198},
      archivePrefix={arXiv},
      primaryClass={cs.RO},
  booktitle={Robotics: Science and Systems (RSS)},
  doi={10.15607/RSS.2025.XXI.106}
}

@inproceedings{omnih2o2024,
  author    = {Tairan He and Zhengyi Luo and Xialin He and Wenli Xiao and Chong Zhang and Weinan Zhang and Kris Kitani and Changliu Liu and Guanya Shi},
  title     = {{OmniH2O}: Universal and Dexterous Human-to-Humanoid Whole-Body Teleoperation and Learning},
  booktitle = {Conference on Robot Learning (CoRL)},
  year      = {2024}
}

@inproceedings{hover2025,
  author    = {Tairan He and Wenli Xiao and Toru Lin and Zhengyi Luo and Zhenjia Xu and Zhenyu Jiang and Jan Kautz and Changliu Liu and Guanya Shi and Xiaolong Wang and Linxi Fan and Yuke Zhu},
  title     = {{HOVER}: Versatile Neural Whole-Body Controller for Humanoid Robots},
  booktitle = {IEEE International Conference on Robotics and Automation (ICRA)},
  year      = {2025}
}

@inproceedings{exbody2024,
  author    = {Xuxin Cheng and Yandong Ji and Junming Chen and Ruihan Yang and Ge Yang and Xiaolong Wang},
  title     = {Expressive Whole-Body Control for Humanoid Robots},
  booktitle = {Robotics: Science and Systems (RSS)},
  year      = {2024}
}

@inproceedings{humanplus2024,
  author    = {Zipeng Fu and Qingqing Zhao and Qi Wu and Gordon Wetzstein and Chelsea Finn},
  title     = {{HumanPlus}: Humanoid Shadowing and Imitation from Humans},
  booktitle = {Conference on Robot Learning (CoRL)},
  year      = {2024}
}

@inproceedings{dao2024boxlocomanipulation,
  author       = {Dao, Jeremy and Duan, Helei and Fern, Alan},
  title        = {Sim-to-Real Learning for Humanoid Box Loco-Manipulation},
  booktitle    = {2024 IEEE International Conference on Robotics and Automation (ICRA)},
  pages        = {16930--16936},
  year         = {2024},
  organization = {IEEE}
}

@inproceedings{amo,
  author       = {Li, Jialong and Cheng, Xuxin and Huang, Tianshu and Yang, Shiqi and Qiu, Ri-Zhao and Wang, Xiaolong},
  title        = {{AMO}: Adaptive Motion Optimization for Hyper-Dexterous Humanoid Whole-Body Control},
  booktitle    = {Proceedings of Robotics: Science and Systems (RSS)},
  address      = {Los Angeles, CA, USA},
  month        = jun,
  year         = {2025},
  doi          = {10.15607/RSS.2025.XXI.061}
}

@inproceedings{he2024learningh20,
  author       = {He, Tairan and Luo, Zhengyi and Xiao, Wenli and Zhang, Chong and Kitani, Kris and Liu, Changliu and Shi, Guanya},
  title        = {Learning Human-to-Humanoid Real-Time Whole-Body Teleoperation},
  booktitle    = {2024 IEEE/RSJ International Conference on Intelligent Robots and Systems (IROS)},
  pages        = {8944--8951},
  year         = {2024},
  organization = {IEEE}
}

@article{26-toussaint-CSRL,
  title         = {Combined Constrained Sampling and Reinforcement Learning for Robotic Manipulation},
  author        = {Toussaint, Marc and Braun, Cornelius V. and Jordana, Armand and Auddy, Sayantan and Cobo-Briesewitz, Eckart and Shcherba, Denis and Burghoff, Tilman and Carpentier, Justin},
  year          = {2026},
  eprint        = {2602.08557},
  archivePrefix = {arXiv},
  primaryClass  = {cs.RO},
  journal={arXiv preprint arXiv:2602.08557}
}

@inproceedings{robopanoptes,
    AUTHOR    = {Xiaomeng Xu AND Dominik Bauer AND Shuran Song}, 
    TITLE     = {{RoboPanoptes}: The All-Seeing Robot with Whole-body Dexterity}, 
    BOOKTITLE = {Proceedings of Robotics: Science and Systems}, 
    YEAR      = {2025}, 
    ADDRESS   = {Los Angeles, CA, USA}, 
    MONTH     = {June}, 
    DOI       = {10.15607/RSS.2025.XXI.042} 
}

@article{blinddexterity,
      title={Blind Dexterity: Whole-Body Humanoid Manipulation via Pure Proprioception}, 
      author={Aditya Bhatt and Oleg Kaidanov and Puze Liu and Jan Peters},
      year={2026},
      eprint={2608.29487},
      archivePrefix={arXiv},
      primaryClass={cs.RO},
  journal={arXiv preprint arXiv:2608.29487}
}

@article{handroid,
  title        = {{Handroid}: Bridging Dexterous Hand and Humanoid},
  author       = {Li, Ruogu and Ma, Chenyang and Li, Sikai and Wei, Zhenyu and Yao, Yunchao and Shi, Haochen and Liu, C. Karen and Song, Shuran and Ding, Mingyu},
  year          = {2026},
  eprint        = {2607.16187},
  archivePrefix = {arXiv},
  primaryClass  = {cs.RO},
  journal={arXiv preprint arXiv:2607.16187}
}

@inproceedings{embracingbulky,
      title={Embracing Bulky Objects with Humanoid Robots: Whole-Body Manipulation with Reinforcement Learning},
      author={Chunxin Zheng and Kai Chen and Zhihai Bi and Yulin Li and Liang Pan and Jinni Zhou and Haoang Li and Jun Ma},
      booktitle={IEEE International Conference on Robotics and Automation (ICRA)},
      pages={6089--6096},
      year={2026},
      doi={10.1109/ICRA57385.2026.11697242}
}

@article{softtorso,
      title={Zero-shot Whole-Body Manipulation with a Large-Scale Soft Robotic Torso via Guided Reinforcement Learning}, 
      author={Curtis C. Johnson and Carlo Alessi and Egidio Falotico and Marc D. Killpack},
      year={2025},
      eprint={2509.23556},
      archivePrefix={arXiv},
      primaryClass={cs.RO},
  journal={arXiv preprint arXiv:2509.23556}
}

@article{tact,
   title={{TACT}: Humanoid Whole-Body Contact Manipulation Through Deep Imitation Learning With Tactile Modality},
   volume={10},
   DOI={10.1109/lra.2025.3580329},
   number={8},
   journal={IEEE Robotics and Automation Letters},
   publisher={Institute of Electrical and Electronics Engineers (IEEE)},
   author={Murooka, Masaki and Hoshi, Takahiro and Fukumitsu, Kensuke and Masuda, Shimpei and Hamze, Marwan and Sasaki, Tomoya and Morisawa, Mitsuharu and Yoshida, Eiichi},
   year={2025},
   month=Aug, pages={7819--7826} }

@inproceedings{explicitcontact,
   title={Explicit Contact Optimization in Whole-Body Contact-Rich Manipulation},
   DOI={10.1109/humanoids58906.2024.10769591},
   booktitle={2024 IEEE-RAS 23rd International Conference on Humanoid Robots (Humanoids)},
   publisher={IEEE},
   author={Levé, Victor and Moura, João and Saito, Namiko and Tonneau, Steve and Vijayakumar, Sethu},
   year={2024},
   month=Nov, pages={451–458} }

@article{posturegen,
   title={Optimization-Based Posture Generation for Whole-Body Contact Motion by Contact Point Search on the Body Surface},
   volume={5},
   ISSN={2377-3774},
   DOI={10.1109/lra.2020.2974689},
   number={2},
   journal={IEEE Robotics and Automation Letters},
   publisher={Institute of Electrical and Electronics Engineers (IEEE)},
   author={Murooka, Masaki and Okada, Kei and Inaba, Masayuki},
   year={2020},
   month=Apr, pages={2905–2912} }

@inproceedings{murooka2015,
  title     = {Whole-body holding manipulation by humanoid robot based on transition graph of object motion and contact},
  author    = {Murooka, Masaki and Inagaki, Yuto and Ueda, Ryohei and Nozawa, Shunichi and Kakiuchi, Yohei and Okada, Kei and Inaba, Masayuki},
  booktitle = {IEEE/RSJ International Conference on Intelligent Robots and Systems (IROS)},
  year      = {2015},
  doi       = {10.1109/IROS.2015.7353933},
  pages     = {3950--3955}
}

@article{contactmimic,
  title   = {{ContactMimic}: Humanoid Object Interaction via Contact Control},
  author  = {Li, Xinyao and He, Xialin and Dong, Runpei and Gupta, Saurabh},
  journal = {arXiv preprint arXiv:2607.08742},
  year    = {2026}
}

@misc{bd_atlas,
  title        = {Training a Humanoid Robot for Hard Work},
  author       = {{Boston Dynamics}},
  howpublished = {\url{https://bostondynamics.com/blog/training-a-humanoid-robot-for-hard-work/}},
  year         = {2026},
}

@misc{figure_helix,
  title        = {Helix 02: Living Room Tidy},
  author       = {{Figure AI}},
  howpublished = {\url{https://www.figure.ai/news/helix-02-living-room-tidy}},
  month        = mar,
  year         = {2026},
}

@inproceedings{ultra,
      title={{ULTRA}: Unified Multimodal Control for Autonomous Humanoid Whole-Body Loco-Manipulation}, 
      author={Xialin He and Sirui Xu and Xinyao Li and Runpei Dong and Liuyu Bian and Yu-Xiong Wang and Liang-Yan Gui},
      year={2026},
      eprint={2603.03279},
      archivePrefix={arXiv},
      primaryClass={cs.RO},
  booktitle={IEEE/RSJ International Conference on Intelligent Robots and Systems (IROS)}
}

@inproceedings{falcon,
    title={{FALCON}: Learning Force-Adaptive Humanoid Loco-Manipulation},
      author={Zhang, Yuanhang and Yuan, Yifu and Gurunath, Prajwal and Gupta, Ishita and Omidshafiei, Shayegan and Agha-mohammadi, Ali-akbar and Vazquez-Chanlatte, Marcell and Pedersen, Liam and He, Tairan and Shi, Guanya},
      year={2026},
  booktitle={Proceedings of the 8th Annual Learning for Dynamics and Control Conference (L4DC)},
  series={PMLR},
  volume={331},
  pages={265--281}
}

@article{grail,
      title={{GRAIL}: Generating Humanoid Loco-Manipulation from {3D} Assets and Video Priors}, 
      author={Tianyi Xie and Haotian Zhang and Jinhyung Park and Zi Wang and Bowen Wen and Jiefeng Li and Xueting Li and Qingwei Ben and Haoyang Weng and Yufei Ye and David Minor and Tingwu Wang and Chenfanfu Jiang and Sanja Fidler and Jan Kautz and Linxi Fan and Yuke Zhu and Zhengyi Luo and Umar Iqbal and Ye Yuan},
      year={2026},
      eprint={2606.05160},
      archivePrefix={arXiv},
      primaryClass={cs.RO},
  journal={arXiv preprint arXiv:2606.05160}
}

@article{motiondisco,
      title={{MotionDisco}: Motion Discovery for Extreme Humanoid Loco-Manipulation}, 
      author={Ilyass Taouil and Michal Ciebielski and Shafeef Omar and Haizhou Zhao and Angela Dai and Aaron M. Johnson and Majid Khadiv},
      year={2026},
      eprint={2606.06139},
      archivePrefix={arXiv},
      primaryClass={cs.RO},
  journal={arXiv preprint arXiv:2606.06139}
}

@inproceedings{omniretarget,
      title={{OmniRetarget}: Interaction-Preserving Data Generation for Humanoid Whole-Body Loco-Manipulation and Scene Interaction}, 
      author={Lujie Yang and Xiaoyu Huang and Zhen Wu and Angjoo Kanazawa and Pieter Abbeel and Carmelo Sferrazza and C. Karen Liu and Rocky Duan and Guanya Shi},
      year={2026},
      eprint={2509.26633},
      archivePrefix={arXiv},
      primaryClass={cs.RO},
  booktitle={IEEE International Conference on Robotics and Automation (ICRA)}
}

@inproceedings{araujo2025retargeting,
  title={Retargeting Matters: General Motion Retargeting for Humanoid Motion Tracking},
  author={Ara{\'u}jo, Jo{\~a}o Pedro and Ze, Yanjie and Xu, Pei and Wu, Jiajun and Liu, C. Karen},
  booktitle={IEEE International Conference on Robotics and Automation (ICRA)},
  pages={6097--6104},
  year={2026},
  doi={10.1109/ICRA57385.2026.11697263}
}

@inproceedings{twist,
    title={{TWIST}: Teleoperated Whole-Body Imitation System},
    author= {Yanjie Ze and Zixuan Chen and João Pedro Araújo and Zi-Ang Cao and Xue Bin Peng and Jiajun Wu and C. Karen Liu},
    booktitle={Conference on Robot Learning (CoRL)},
    year= {2025}
}

@article{dactyl,
  title={Learning dexterous in-hand manipulation},
  author={{OpenAI} and Andrychowicz, Marcin and Baker, Bowen and Chociej, Maciek and J{\'o}zefowicz, Rafal and McGrew, Bob and Pachocki, Jakub and Petron, Arthur and Plappert, Matthias and Powell, Glenn and Ray, Alex and others},
  journal={The International Journal of Robotics Research},
  volume={39},
  number={1},
  pages={3--20},
  year={2020},
  publisher={SAGE Publications Sage UK: London, England}
}

@inproceedings{hora,
  title={In-hand object rotation via rapid motor adaptation},
  author={Qi, Haozhi and Kumar, Ashish and Calandra, Roberto and Ma, Yi and Malik, Jitendra},
  booktitle={Conference on Robot Learning},
  series={PMLR},
  volume={205},
  pages={1722--1732},
  year={2023},
  organization={PMLR}
}

@article{visualdexterity,
  title={Visual dexterity: In-hand reorientation of novel and complex object shapes},
  author={Chen, Tao and Tippur, Megha and Wu, Siyang and Kumar, Vikash and Adelson, Edward and Agrawal, Pulkit},
  journal={Science Robotics},
  volume={8},
  number={84},
  pages={eadc9244},
  year={2023},
  publisher={American Association for the Advancement of Science}
}

@inproceedings{intermimic,
  title={{InterMimic}: Towards Universal Whole-Body Control for Physics-Based Human-Object Interactions},
  author={Xu, Sirui and Ling, Hung Yu and Wang, Yu-Xiong and Gui, Liang-Yan},
  booktitle={2025 IEEE/CVF Conference on Computer Vision and Pattern Recognition (CVPR)},
  pages={12266--12277},
  year={2025},
  organization={IEEE}
}

@article{liu2022synthesizingdiversephysicallystable,
      title={Synthesizing Diverse and Physically Stable Grasps with Arbitrary Hand Structures using Differentiable Force Closure Estimator}, 
      author={Tengyu Liu and Zeyu Liu and Ziyuan Jiao and Yixin Zhu and Song-Chun Zhu},
      year={2022},
      eprint={2104.09194},
      archivePrefix={arXiv},
      primaryClass={cs.RO},
  journal={IEEE Robotics and Automation Letters},
  volume={7},
  number={1},
  pages={470--477},
  doi={10.1109/LRA.2021.3129138}
}

@inproceedings{wang2023dexgraspnetlargescaleroboticdexterous,
      title={{DexGraspNet}: A Large-Scale Robotic Dexterous Grasp Dataset for General Objects Based on Simulation},
      author={Ruicheng Wang and Jialiang Zhang and Jiayi Chen and Yinzhen Xu and Puhao Li and Tengyu Liu and He Wang},
      booktitle={IEEE International Conference on Robotics and Automation (ICRA)},
      pages={11359--11366},
      year={2023},
      doi={10.1109/ICRA48891.2023.10160982}
}

@inproceedings{chen2025bodexscalableefficientrobotic,
      title={{BODex}: Scalable and Efficient Robotic Dexterous Grasp Synthesis Using Bilevel Optimization},
      author={Jiayi Chen and Yubin Ke and He Wang},
      booktitle={IEEE International Conference on Robotics and Automation (ICRA)},
      year={2025},
      doi={10.1109/ICRA55743.2025.11127930}
}

@misc{2025foamtoolsphericalapproximation,
      title={Foam: A Tool for Spherical Approximation of Robot Geometry},
      author={Sai Coumar and Gilbert Chang and Nihar Kodkani and Zachary Kingston},
      year={2025},
      eprint={2503.13704},
      archivePrefix={arXiv},
      primaryClass={cs.RO},
      url={https://arxiv.org/abs/2503.13704},
}

@article{schulman2017ppo,
  title={Proximal Policy Optimization Algorithms},
  author={Schulman, John and Wolski, Filip and Dhariwal, Prafulla and Radford, Alec and Klimov, Oleg},
  year={2017},
  eprint={1707.06347},
  archivePrefix={arXiv},
  primaryClass={cs.LG},
  journal={arXiv preprint arXiv:1707.06347}
}

@inproceedings{twist2,
  title={{TWIST2}: Scalable, Portable, and Holistic Humanoid Data Collection System},
  author={Ze, Yanjie and Zhao, Siheng and Wang, Weizhuo and Kanazawa, Angjoo and Duan, Rocky and Abbeel, Pieter and Shi, Guanya and Wu, Jiajun and Liu, C. Karen},
  booktitle={IEEE International Conference on Robotics and Automation (ICRA)},
  pages={15900--15907},
  year={2026},
  doi={10.1109/ICRA57385.2026.11696896}
}

@article{humdex,
  title={{HumDex}: Humanoid Dexterous Manipulation Made Easy},
  author={Heng, Liang and Tang, Yihe and Xu, Jiajun and Bao, Henghui and Huang, Di and Wang, Yue},
  journal={arXiv preprint arXiv:2603.12260},
  year={2026}
}

@article{weave,
  title={{Weave}: Learning Whole-Body Dexterous Loco-Manipulation from Human-Object Interactions},
  author={Cao, Liu and Wu, Xingze and Cui, Jingzhi and Xu, Botian and Pei, Mingzhi and Chen, Ruoqu and Xu, Mengdi},
  journal={arXiv preprint arXiv:2609.16683},
  year={2026}
}

@misc{resmimic,
  title         = {{ResMimic}: From General Motion Tracking to Humanoid Whole-body Loco-Manipulation via Residual Learning},
  author        = {Zhao, Siheng and Ze, Yanjie and Wang, Yue and Liu, C. Karen and Abbeel, Pieter and Shi, Guanya and Duan, Rocky},
  year          = {2025},
  eprint        = {2510.05070},
  archivePrefix = {arXiv},
  primaryClass  = {cs.RO},
}

@misc{punyo_pgrl,
  title         = {Plan-Guided Reinforcement Learning for Whole-Body Manipulation},
  author        = {Zhang, Mengchao and Barreiros, Jose and Onol, Aykut Ozgun},
  year          = {2023},
  eprint        = {2310.12263},
  archivePrefix = {arXiv},
  primaryClass  = {cs.RO},
}

@article{punyo_scirobotics,
  title   = {Learning contact-rich whole-body manipulation with example-guided reinforcement learning},
  author  = {Barreiros, Jose A. and {\"O}nol, Aykut {\"O}zg{\"u}n and Zhang, Mengchao and Creasey, Sam and Goncalves, Aimee and Beaulieu, Andrew and Bhat, Aditya and Tsui, Kate M. and Alspach, Alex},
  journal = {Science Robotics},
  volume  = {10},
  number  = {105},
  pages   = {eads6790},
  year    = {2025},
  doi     = {10.1126/scirobotics.ads6790},
}

@misc{zakka2026mjlablightweightframeworkgpuaccelerated,
  title         = {{mjlab}: A Lightweight Framework for {GPU}-Accelerated Robot Learning},
  author        = {Kevin Zakka and Qiayuan Liao and Brent Yi and Louis Le Lay and Koushil Sreenath and Pieter Abbeel},
  year          = {2026},
  eprint        = {2601.22074},
  archivePrefix = {arXiv},
  primaryClass  = {cs.RO},
  url           = {https://arxiv.org/abs/2601.22074},
}

@inproceedings{pinto2018asymmetric,
  author    = {Lerrel Pinto and Marcin Andrychowicz and Peter Welinder and Wojciech Zaremba and Pieter Abbeel},
  title     = {Asymmetric Actor Critic for Image-Based Robot Learning},
  booktitle = {Proceedings of Robotics: Science and Systems},
  year      = {2018},
  address   = {Pittsburgh, Pennsylvania},
  month     = jun,
  doi       = {10.15607/RSS.2018.XIV.008},
}

@article{li2026teledexter,
  title={Towards Human-level Dexterous Teleoperation},
  author={Li, Puhao and Chen, Zeyuan and Wu, Yingying and Wei, Pengkun and Li, Yuyang and Wang, Tianyu and Shi, Jiaxiao and Yu, Mingrui and Jia, Baoxiong and Zhu, Song-Chun and Liu, Tengyu and Huang, Siyuan},
  journal={arXiv preprint arXiv:2607.11481},
  year={2026}
}
